\documentclass{IEEEtran}
\usepackage{authblk}
\usepackage{amsfonts}
\usepackage{cite}
\usepackage{amsmath}
\usepackage{amssymb}
\usepackage{amsthm}
\usepackage{graphicx}
\usepackage{caption}
\usepackage[colorlinks]{hyperref} 
\usepackage[normalem]{ulem} 
\usepackage[dvipsnames]{xcolor} 
\usepackage{subfigure}
\usepackage{algorithmic,algorithm}
\usepackage{booktabs}
\usepackage{epstopdf}
\usepackage{balance}

\graphicspath{{./images/}}

\theoremstyle{plain}
\newtheorem{theorem}{Theorem}
\newtheorem{lemma}{Lemma}

\theoremstyle{definition}
\newtheorem{definition}{Definition}
\newtheorem{assumption}{Assumption}

\newenvironment{assumptionprime}[1]
  {\renewcommand{\theassumption}{\ref{#1}$'$}%
   \addtocounter{assumption}{-1}%
   \begin{assumption}}
  {\end{assumption}}

\theoremstyle{remark}
\newtheorem{remark}{Remark}

\newcommand{\ba}        {\mathbf{a}}
\newcommand{\bA}        {\mathbf{A}}

\newcommand{\B}         {\mathbb{B}}

\newcommand{\bB}        {\mathbf{B}}
\newcommand{\cB}        {\mathcal{B}}

\newcommand{\C}         {\mathbb{C}}

\newcommand{\bc}        {\mathbf{c}}

\newcommand{\cC}        {\mathcal{C}}

\newcommand{\E}         {\mathbb{E}}

\newcommand{\be}        {\mathbf{e}}
\newcommand{\bE}        {\mathbf{E}}
\newcommand{\cE}        {\mathcal{E}}

\newcommand{\cF}        {\mathcal{F}}

\newcommand{\bg}        {\mathbf{g}}

\newcommand{\cG}        {\mathcal{G}}

\newcommand{\wtg}       {\widetilde{g}}

\newcommand{\bh}        {\mathbf{h}}

\newcommand{\bI}        {\mathbf{I}}

\newcommand{\cJ}        {\mathcal{J}}

\newcommand{\cN}        {\mathcal{N}}

\newcommand{\cO}        {\mathcal{O}}

\newcommand{\bbP}       {\mathbb{P}}

\newcommand{\bq}        {\mathbf{q}}

\newcommand{\R}         {\mathbb{R}}

\newcommand{\cR}        {\mathcal{R}}

\newcommand{\cS}        {\mathcal{S}}

\newcommand{\bu}        {\mathbf{u}}

\newcommand{\V}         {\mathbb{V}}

\newcommand{\bv}        {\mathbf{v}}

\newcommand{\bw}        {\mathbf{w}}

\newcommand{\cW}        {\mathcal{W}}

\newcommand{\whbw}      {\widehat{\bw}}

\newcommand{\bx}        {\mathbf{x}}

\newcommand{\cX}        {\mathcal{X}}

\newcommand{\by}        {\mathbf{y}}
\newcommand{\bY}        {\mathbf{Y}}

\newcommand{\bz}        {\mathbf{z}}

\newcommand{\cZ}        {\mathcal{Z}}

\newcommand{\balpha}    {\boldsymbol{\alpha}}

\newcommand{\bPhi}      {\boldsymbol{\Phi}}

\newcommand{\bOmega}    {\boldsymbol{\Omega}}

\newcommand{\bone}      {\mathbf{1}}

\DeclareMathOperator*{\argmin}  {arg\,min}
\DeclareMathOperator*{\argmax}  {arg\,max}

\newcommand{\ERM}       {\textsf{\textsc{erm}}}

\newcommand{\revise}[1]{#1}

\begin{document}

\title{BRIDGE: Byzantine-resilient Decentralized Gradient Descent}

\author{Cheng Fang, Zhixiong Yang, and Waheed U.\ Bajwa%
\thanks{C.\ Fang ({\tt cf446@soe.rutgers.edu}) and W.U.\ Bajwa ({\tt waheed.bajwa@rutgers.edu}) are with the Department of Electrical and Computer Engineering, Rutgers University--New Brunswick, NJ 08854. Z.\ Yang ({\tt zhixiong.yang@bluedanube.com}) completed this work as part of his PhD dissertation at Rutgers University; he is now a Machine Learning Systems Engineer at Blue Danube Systems.}%
\thanks{This work is supported in part by the National Science Foundation under awards CCF-1453073, CCF-1907658, and OAC-1940074, and by the Army Research Office under award W911NF2110301.}}

\maketitle

\begin{abstract}
Machine learning has begun to play a central role in many applications. A multitude of these applications typically also involve datasets that are distributed across multiple computing devices/machines due to either design constraints (e.g., multi-agent and Internet-of-Things systems) or computational/privacy reasons (e.g., large-scale machine learning on smartphone data). Such applications often require the learning tasks to be carried out in a decentralized fashion, in which there is no central server that is directly connected to all nodes. In real-world decentralized settings, nodes are prone to undetected failures due to malfunctioning equipment, cyberattacks, etc., which are likely to crash non-robust learning algorithms. The focus of this paper is on robustification of decentralized learning in the presence of nodes that have undergone Byzantine failures. The Byzantine failure model allows faulty nodes to arbitrarily deviate from their intended behaviors, thereby ensuring designs of the most robust of algorithms. But the study of Byzantine resilience within decentralized learning, in contrast to distributed learning, is still in its infancy. In particular, existing Byzantine-resilient decentralized learning methods either do not scale well to large-scale machine learning models, or they lack statistical convergence guarantees that help characterize their generalization errors. In this paper, a scalable, Byzantine-resilient decentralized machine learning framework termed \textbf{B}yzantine-\textbf{r}es\textbf{i}lient \textbf{d}ecentralized \textbf{g}radient d\textbf{e}scent (BRIDGE) is introduced. Algorithmic and statistical convergence guarantees for one variant of BRIDGE are also provided in the paper for both strongly convex problems and a class of nonconvex problems. In addition, large-scale decentralized learning experiments are used to establish that the BRIDGE framework is scalable and it delivers competitive results for Byzantine-resilient convex and nonconvex learning.
\end{abstract}

\section{Introduction}\label{section introduction}
One of the fundamental tasks of \emph{machine learning} (ML) is to learn a model using training data that minimizes the statistical risk~\cite{Vapnik2013nature}. A typical technique that accomplishes this task is \emph{empirical risk minimization} (ERM) of a loss function~\cite{sebastiani2002machine,kotsiantis2007supervised,bengio2009learning,Mohri2012foundations,MLtextbook}. Under the ERM framework, an ML model is learned by an optimization algorithm that tries to minimize the average loss with respect to the training data that are assumed available at a single location. In many recent applications of ML, however, training data tend to be geographically distributed; examples include the multi-agent and Internet-of-Things systems, smart grids, sensor networks, etc. In several other recent applications of ML, the training data cannot be gathered at a single machine due to either the massive scale of data and/or privacy concerns; examples in this case include the social network data, smartphone data, healthcare data, etc. The applications in both such cases require that the ML model be learned using training data that are distributed over a network. When the ML/optimization algorithm in such applications requires a central coordinating server connected to all the nodes in the network, the resulting framework is often referred to as \emph{distributed learning}~\cite{YangGangEtAl.ISPM20}. Practical constraints many times also require an application to accomplish the learning tasks without a central server~\cite{nokleby2020}, in which case the resulting framework is referred to as \emph{decentralized learning}.

The focus of this paper is on decentralized learning\revise{, with a particular emphasis on characterizing the \emph{sample complexity} of the decentralized learning algorithm---i.e., the rate, as a function of the number of training data samples, at which the ERM solution approaches the \emph{Bayes optimal solution} in a decentralized setting~\cite{Mohri2012foundations,nokleby2020}}. While decentralized learning has a rich history
, a significant fraction of that work has focused on the faultless setting~\cite{predd2006distributed,boyd2011distributed,AliH.Sayed2014,nedic2018,sun2021decentralized}. But real-world decentralized systems are bound to undergo failures because of malfunctioning equipment, cyberattacks, and so on~\cite{Driscoll2003byzantine}. And when the failures happen and go undetected, the learning algorithms designed for the faultless networks break down\cite{Su2016fault,YangGangEtAl.ISPM20}. Among the different types of failures in the network, the so-called \emph{Byzantine failure}~\cite{Lamport1982byzantine} is considered the most general, as it allows the faulty/compromised nodes to arbitrarily deviate from the agreed-upon protocol~\cite{Driscoll2003byzantine}. Byzantine failures are the hardest to safeguard against and can easily jeopardize the ability of the network to reach consensus~\cite{dutta2005best,sousa2012byzantine}. Moreover,  it has been shown in~\cite{Su2016fault} that a single Byzantine node with a simple strategy can lead to the failure of a decentralized learning algorithm. The overarching goal of this paper is to develop \revise{and (algorithmically and statistically) analyze} an efficient decentralized learning algorithm that is provably resilient against Byzantine failures in decentralized settings with respect to both convex and nonconvex loss functions.

\subsection{Relationship to prior works}\label{ssec:prior.work}
Although the model of Byzantine failure was brought up decades ago, it has attracted the attention of ML researchers only very recently. Motivated by applications in large-scale machine learning~\cite{nokleby2020}, much of that work has focused solely on the distributed learning setup such as the parameter--server setting~\cite{Muli2014} and the federated learning setting~\cite{jakub2016}. A necessarily incomplete list of these works, most of which have developed and analyzed Byzantine-resilient distributed learning approaches from the perspective of stochastic gradient descent, include \cite{blanchard2017machine,draco2017chen,cao2018robust,su2018securing,yin2018defending,damaskinos2018asynchronous,mhamdi2018hidden,yin2018byzantine,alistarh2018byzantine,xie2018zeno,xie2018generalized,xie2018phocas,chen2019distributed,rajput2019detox,li2018rsa,jin2019distributed,lin2019byzantine,ghosh2019robust,data2019data,el2019sgd,xie2019zeno++,elmhamdi2019fast}.  Nonetheless, translating the algorithmic and analytical insights from the distributed learning setups to the decentralized ones, which lack central coordinating servers, is a nontrivial endeavor. As such, despite the plethora of work on Byzantine-resilient distributed learning, the problem of Byzantine-resilient decentralized learning---with the exception of a handful of works discussed in the following---largely remains unexplored in the literature.

In terms of decentralized learning in general, there are three broad classes of iterative algorithms that can be utilized for decentralized training purposes. The first one of these classes of algorithms corresponds to first-order methods such as the \emph{distributed gradient descent} (DGD) and its (stochastic) variants~\cite{nedic2009,ram2010distributed, nedic2015distributed,nedic2020}. The iterative methods in this class have low (local) computational complexity, which makes them particularly well suited for large-scale problems. The second class of algorithms involves the use of augmented Lagrangian-based methods~\cite{forero2010consensus,mota2013admm,shi2014on}, which require each node in the network to locally solve an optimization subproblem. The third class of algorithms includes second-order methods\cite{Mokhtari2016decentralized, Mokhtari2017network}, which typically have high computational and/or communications cost. Although the decentralized learning methods within these three classes of algorithms have their own sets of strengths and weaknesses, all of these traditional works assume faultless operations within the decentralized network.

Within the context of Byzantine failures in decentralized systems, some of the first works focused on the problem of Byzantine-resilient averaging consensus~\cite{leblanc2013resilient,vaidya2014iterative}. These works were then leveraged to develop theory and algorithms for Byzantine-resilient decentralized learning for the case of scalar-valued models~\cite{Su2016fault,sundaram2018distributed}. \revise{But neither of these works are applicable to the general vector-valued ML framework being considered in this paper.} In parallel, some researchers have also developed Byzantine-resilient decentralized learning methods for some specific vector-valued problems that include the decentralized support vector machine~\cite{yang2016rdsvm} and decentralized estimation~\cite{xu2018robust,mitra2019resilient,su2018finite}.

Similar to the classical ML framework, however, there is a need to develop \revise{and algorithmically/statistically analyze} Byzantine-resilient decentralized learning methods for vector-valued models for general---rather than specialized---loss functions, which can be broadly divided into two classes of convex and nonconvex loss functions. The first work in the literature that tackled this problem is \cite{yang2019byrdie}, which developed a decentralized coordinate-descent-based learning algorithm termed ByRDiE and established its resilience to Byzantine failures in the case of a loss function that is given by the sum of a convex differentiable function and a strictly convex and smooth regularizer. The analysis in \cite{yang2019byrdie} also provided rates for algorithmic convergence as well as statistical convergence (i.e., sample complexity) of ByRDiE. One of the limitations of \cite{yang2019byrdie} is its exclusive focus on convex loss functions for the purposes of analysis. More importantly, however, the coordinate-descent nature of ByRDiE makes it slow and inefficient for learning of large-scale models. Let $d$ denote the number of parameters in the ML model being trained (e.g., the number of weights in a deep neural network). One iteration of ByRDiE then requires updating the $d$ coordinates of the model in $d$ network-wide collaborative steps, each one of which requires a computation of the local $d$-dimensional gradient at each node in the network. In the case of large-scale models such as the deep neural networks with tens or hundreds of thousands of parameters, the local computation costs as well as the network-wide coordination and communications overhead of such an approach can be prohibitive for many applications. \revise{By contrast, since the algorithmic developments in this paper are based on the gradient-descent method, the resulting computational framework is highly efficient and scalable in a decentralized setting. And while the algorithmic and statistical convergence results derived in here match those for ByRDiE in the case of convex loss functions, the proposed framework is fundamentally different from ByRDiE and therefore necessitates its own theoretical analysis.}

We conclude by noting that some additional works \cite{Kuwaranancharoen2020ByzantineResilientDO,Peng2020ByzantineRobustDS,Guo2020TowardsBL,ElMhamdi2020CollaborativeLA} relevant to the topic of Byzantine-resilient decentralized learning have appeared during the course of revising this paper, which are being discussed here for the sake of completeness. \revise{It is worth reminding the reader, however, that the work in this paper predates these recent efforts. Equally important, none of these works provide statistical convergence rates for the proposed methods. Additionally, the work in \cite{Kuwaranancharoen2020ByzantineResilientDO} only focuses on convex loss functions and it does not provide any convergence rates. Further, the ability of the proposed algorithm to defend against a large number of Byzantine nodes severely diminishes with an increase in the problem dimension.} In contrast, the authors in \cite{Peng2020ByzantineRobustDS} focus on Byzantine-resilient decentralized learning in the presence of non-uniformly distributed data and time-varying networks. The focus in this work is also only on convex loss functions and the performance of the proposed algorithm is worse than that of the approach advocated in this work for static networks and uniformly distributed data. Next, an algorithm termed MOZI is proposed in \cite{Guo2020TowardsBL}, with the focus once again being on convex loss functions. The resilience of MOZI, however, requires an aggressive two-step `filtering' operation, which limits the maximum number of Byzantine nodes that can be handled by the algorithm. The analysis in \cite{Guo2020TowardsBL} also makes the unrealistic assumption that the faulty nodes always send messages that are `outliers' relative to those of the regular nodes. Finally, the only paper in the literature that has investigated Byzantine-resilient decentralized learning for nonconvex loss functions is \cite{ElMhamdi2020CollaborativeLA}. The authors in this work have introduced three methods, among which the so-called ICwTM method is effectively a variant of our approach. The ICwTM algorithm, however, has at least twice the communications overhead of our approach, since it requires the neighbors to exchange both their local models and local gradients. In addition, \cite{ElMhamdi2020CollaborativeLA} requires the nodes to have the same initialization and it does not bring out the dependence of the network topology on the learning problem.

\begin{table*}[t]
\centering
\begin{tabular}{|c c c c c|}
 \hline
 {\bf Algorithm} & {\bf Nonconvex} & {\bf Byzantine failures} & {\bf Algorithmic convergence rate} & {\bf Statistical convergence rate}\\
 \hline\hline
 ByRDiE~\cite{yang2019byrdie} &$\times$ & $\surd$ & $\surd$ & $\surd$\\
 Kuwaranancharoen et.\ al~\cite{Kuwaranancharoen2020ByzantineResilientDO} & $\times$ & $\surd$ & $\times$ & $\times$ \\
 Peng and Ling~\cite{Peng2020ByzantineRobustDS} & $\times$ & $\surd$ & $\surd$ & $\times$ \\
 MOZI~\cite{Guo2020TowardsBL} & $\times$ & $\surd$ & $\surd$ & $\times$\\
 ICwTM~\cite{ElMhamdi2020CollaborativeLA} &$\surd$ & $\surd$ & $\surd$ & $\times$\\
 DGD~\cite{nedic2015distributed} & $\times$ & $\times$ & $\surd$ & $\times$ \\
 NEXT~\cite{Lorenzo2016NEXTIN} & $\surd$ & $\times$ & $\times$ & $\times$\\
 Nonconvex DGD~\cite{Zeng2018OnND} & $\surd$ & $\times$ & $\surd$ & $\times$ \\
 D-GET~\cite{Sun2019ImprovingTS} & $\surd$ & $\times$ & $\surd$ & $\surd$ \\
 GT-SARAH~\cite{xin2021fast} & $\surd$ & $\times$ & $\surd$ & $\surd$ \\
 {\bf BRIDGE (This paper)} & $\surd$ & $\surd$ & $\surd$ & $\surd$\\
 \hline
\end{tabular}
\caption{Comparison of BRIDGE with different \revise{vector-valued} decentralized learning/optimization methods in the literature.}
\label{table:intro.comparison}
\end{table*}

\revise{
\begin{remark}\label{rem:Lie2022Byzantine}
While this paper was in review, a related work~\cite{Lie2022Byzantine} appeared on a preprint server; this recent work on Byzantine-resilient decentralized learning, in contrast to our paper, studies a more general class of nonconvex loss functions and also allows the distribution of training data at the regular nodes to be heterogeneous. However, in addition to the fact that~\cite{Lie2022Byzantine} significantly postdates our work, the main result in \cite{Lie2022Byzantine} relies on clairvoyant knowledge of several network-wide parameters, including the subset of Byzantine nodes within the network, and also requires the maximum `cumulative mixing weight' associated with the Byzantine nodes to be impractically small (e.g., even in the case of a \emph{fully connected} network, the cumulative weight must be no greater than $9.76 \times 10^{-5}$).
\end{remark}
}

\subsection{Our contributions}\label{ssec:our.contributions}
One of the main contributions of this paper is the introduction of an efficient and scalable algorithmic framework for Byzantine-resilient decentralized learning. The proposed framework, termed \emph{\textbf{B}yzantine-\textbf{r}es\textbf{i}lient \textbf{d}ecentralized \textbf{g}radient d\textbf{e}scent} (BRIDGE), overcomes the computational and communications overhead associated with the one-coordinate-at-a-time update pattern of ByRDiE through its use of the gradient descent-style updates. Specifically, the network nodes locally compute the $d$-dimensional gradient (and exchange the local $d$-dimensional model) only once in each iteration of the BRIDGE framework, as opposed to the $d$ computations of the $d$-dimensional gradient in each iteration of ByRDiE. The BRIDGE framework therefore has significantly less local computational cost due to fewer gradient computations, and it also has smaller network-wide coordination and communications overhead due to fewer exchange of node-to-node messages. Note that BRIDGE is being referred to as a framework since it allows for multiple variants of a single algorithm depending on the choice of the \emph{screening} method used within the algorithm for resilience purposes; see Section \ref{section BRIDGE} for further details.
%

Another main contribution of this paper is analysis of one of the variants of BRIDGE, termed BRIDGE-T, for resilience against Byzantine failures in the network. The analysis enables us to provide both algorithmic convergence rates and statistical convergence rates for BRIDGE-T for certain classes of convex and nonconvex loss functions, with the rates derived for the convex setting matching those for ByRDiE~\cite{yang2019byrdie}. The final main contribution of this paper is reporting of large-scale numerical results on the MNIST~\cite{Lecun1998} \revise{and CIFAR-10~\cite{krizhevsky2009learning} datasets} for both convex and nonconvex decentralized learning problems in the presence of Byzantine failures. The reported results, which include both \emph{independent and identically distributed} (i.i.d.) and non-i.i.d.\ datasets within the network, highlight the benefits of the BRIDGE framework and validate our theoretical findings.

In summary, and to the best of our knowledge, BRIDGE is the first 
Byzantine-resilient decentralized learning algorithm that is scalable, has results for \revise{a class of} nonconvex learning problems, and provides rates for both algorithmic and statistical convergence. We also refer the reader to Table~\ref{table:intro.comparison}, which compares BRIDGE with recent works in both faultless and faulty \revise{vector-valued} decentralized optimization/learning settings. \revise{Additional relevant works not appearing in this table include \cite{Su2016fault,sundaram2018distributed}, since they limit themselves to scalar-valued problems, and \cite{Lie2022Byzantine}, since it substantially postdates our work. Further, \cite{Su2016fault,sundaram2018distributed} neither study nonconvex loss functions nor derive the statistical convergence rates, while the main result in \cite{Lie2022Byzantine}---despite the generality of its problem setup---is significantly restrictive, as discussed in Remark~\ref{rem:Lie2022Byzantine}.}

\subsection{Notation and organization}
The following notation is used throughout the rest of the paper. We denote scalars with regular-faced lowercase and uppercase letters (e.g., $a$ and $A$), vectors with bold-faced lowercase letters (e.g., $\ba$), and matrices with bold-faced uppercase letters (e.g., $\bA$). All vectors are taken to be column vectors, while $[\ba]_k$ and $[\bA]_{ij}$ denote the $k$-th element of vector $\ba$ and the $(i,j)$-th element of matrix $\bA$, respectively. We use $\| \ba\| $ to denote the $\ell_2$-norm of $\ba$, $\mathbf{1}$ to denote the vector of all ones, and $\bI$ to denote the identity matrix, while $(\cdot)^T$ denotes the transpose operation. Given two matrices $\bA$ and $\bB$, the notation $\bA \succeq \bB$ signifies that $\bA - \bB$ is a positive semidefinite matrix. We also use $\langle \ba_1, \ba_2 \rangle$ to denote the inner product between two vectors. For a given vector $\ba$ and nonnegative constant $\gamma$, we denote the $\ell_2$-ball of radius $\gamma$ centered around $\ba$ as $\B(\ba,\gamma):=\{\ba': \|\ba-\ba'\|\leq \gamma \}$. Finally, given a set, $\vert \cdot\vert$ denotes its cardinality, while we use the notation $\cG(\cJ,\cE)$ to denote a graph with the set of nodes $\cJ$ and edges $\cE$.

\looseness=-1 The rest of this paper is organized as follows.  Section~\ref{section problem formulation} provides a mathematical formulation of the Byzantine-resilient decentralized learning problem, along with a formal definition of a Byzantine node and various assumptions on the loss function. Section~\ref{section BRIDGE} introduces the BRIDGE framework and discusses different variants of the BRIDGE algorithm. Section~\ref{sec: theoretical analysis} provides theoretical guarantees for the BRIDGE-T algorithm for certain classes of convex and nonconvex loss functions, which include guarantees for network-wide consensus among the nonfaulty nodes and statistical convergence. Section~\ref{section numerical analysis} reports results corresponding to numerical experiments on the MNIST \revise{and CIFAR-10 datasets} for both convex and nonconvex learning problems, establishing the usefulness of the BRIDGE framework for Byzantine-resilient decentralized learning. We conclude the paper in Section~\ref{section conclusion}, while the appendices contain the proofs of the main lemmas and theorems.

\section{Problem Formulation}\label{section problem formulation}
\subsection{Preliminaries}
Let $(\bw,\bz) \mapsto f(\bw,\bz)$ be a non-negative-valued (and possibly regularized) \emph{loss function} that maps the tuple of a \emph{model} $\bw$ and a \emph{data sample} $\bz$ to the corresponding loss $f(\bw,\bz)$. Without loss of much generality, we assume the model $\bw$ in this paper to be a parametric one, i.e.,  $\bw \in \R^d$ (e.g., $d$ could be the number of parameters in a deep neural network). The data sample $\bz$, on the other hand, corresponds to a random variable on some probability space $(\Omega,\cF,\bbP)$, i.e., $\bz$ is $\cF$-measurable and has been drawn from the sample space $\Omega$ according to the probability law $\bbP$. The holy grail in machine learning (ML) is to obtain an optimal model $\bw^*$ that minimizes the expected loss, termed the \emph{statistical risk} \cite{Mohri2012foundations,MLtextbook}, i.e.,
\begin{align}
    \label{eqn: SRM.equation}
    \bw^* \in \argmin_{\bw \in \R^d} \E_{\bbP}[f(\bw,\bz)].
\end{align}
A model $\bw^*$ that satisfies \eqref{eqn: SRM.equation} is termed a \emph{statistical risk minimizer} (also known as a \emph{Bayes optimal model}). In the real world, however, one seldom has access to the distribution of $\bz$, which precludes the use of the statistical risk $\E_{\bbP}[f(\bw,\bz)]$ in any computations. Instead, a common approach utilized in ML is to leverage a collection $\cZ := \{\bz_n\}_{n=1}^N$ of data samples that have been drawn according to the law $\bbP$ and solve an empirical variant of \eqref{eqn: SRM.equation} as follows \cite{Mohri2012foundations,MLtextbook}:
\begin{align}
    \label{eqn: central.ERM.equation}
    \bw^*_{\ERM} \in \argmin_{\bw \in \R^d} \frac{1}{N} \sum_{n=1}^N f(\bw,\bz_n).
\end{align}
This approach, which is termed as the \emph{empirical risk minimization} (ERM), typically relies on an optimization algorithm to solve for $\bw^*_{\ERM}$. \revise{The resulting solution $\whbw$, from the perspective of an ML practitioner, must satisfy two criteria: ($i$) it should have fast algorithmic convergence, measured in terms of the algorithmic convergence rate, to a fixed point (often taken to be a stationary point of \eqref{eqn: central.ERM.equation} in centralized settings); and ($ii$) it should have fast statistical convergence, often specified in terms of the sample complexity (number of samples), to a statistical risk minimizer. Our focus in this paper, in contrast to several related prior works (cf.~Table~\ref{table:intro.comparison}), is on both the algorithmic and the statistical convergence of the ERM solution.} The final set of results in this case rely on a number of assumptions on the loss function $f(\bw,\bz)$, stated below.
\begin{assumption}[Bounded and Lipschitz gradients]\label{assumption lipschitz}
The loss function $f(\bw,\bz)$ is differentiable in the first argument $\bbP$-almost surely (a.s.) and the gradient of $f(\bw,\bz)$ with respect to the first argument, denoted as $\nabla f(\bw,\bz)$, is bounded and $L'$-Lipschitz a.s., i.e.,\footnote{Unless specified otherwise, all \emph{almost sure} statements in the paper are to be understood with respect to the probability law $\bbP$.} $\forall \bw  \in \R^d, \|\nabla f(\bw,\bz)\| \leq L$ a.s.\ and $$ \forall \bw_1,\bw_2 \in \R^d, \ \|\nabla f(\bw_1,\bz)-\nabla f(\bw_2,\bz)\| \leq L'\| \bw_1-\bw_2\| \hspace{0.1cm} \text{a.s.}$$
\end{assumption}
In the literature, functions with $L'$-Lipschitz gradients are also referred to as $L'$-smooth functions. Assumption~\ref{assumption lipschitz} implies the loss function is itself a.s.\ $L$-Lipschitz continuous~\cite{sohrab2003basic}, i.e., $\forall \bw_1,\bw_2 \in \R^d, \ | f(\bw_1,\bz)-f(\bw_2,\bz)|\leq L\| \bw_1-\bw_2\|$ a.s.

\begin{assumption}[Bounded training loss]\label{assumption finite value}
The loss function is a.s.\ bounded over the training samples, i.e., there exists a constant $C$ such that $\sup_{\bw \in \R^d,\bz \in \cZ} f(\bw,\bz)\leq C <\infty$ a.s.
\end{assumption}

The analysis carried out in this paper considers two different classes of loss functions, namely, convex functions and nonconvex functions. In the case of analysis for the convex loss functions, we make the following assumption.
\begin{assumption}[Strong convexity]\label{assumption strongly convex}
The loss function $f(\bw,\bz)$ is a.s.\ $\lambda$-strongly convex in the first argument, i.e.,  $$\forall \bw_1,\bw_2 \in \R^d, \ f(\bw_1, \bz)\geq f(\bw_2, \bz)+\langle \nabla f(\bw_2,\bz), \bw_1-\bw_2 \rangle$$
$$\qquad\qquad\qquad\qquad\qquad + \frac{\lambda}{2}\| \bw_1-\bw_2 \|^2 \quad \text{a.s.}$$
\end{assumption}
Note that the Lipschitz gradients assumption can be relaxed to Lipschitz subgradients in the case of strongly convex loss functions. Some examples of loss functions that satisfy Assumptions~\ref{assumption lipschitz} and~\ref{assumption strongly convex} arise in ridge regression, elastic net regression, $\ell_2$-regularized logistic regression, and $\ell_2$-regularized training of support vector machines when the optimization variable $\bw$ is constrained to belong to a bounded set in $\R^d$. Our discussion in the sequel shows that $\bw$ indeed remains bounded for the algorithms in consideration, justifying the usage of Assumptions~\ref{assumption lipschitz} and~\ref{assumption strongly convex}. And while Assumption~\ref{assumption finite value}, as currently stated, would not be satisfied for the aforementioned regularized problems, the analysis in the paper only requires boundedness of the data-dependent term(s) of the loss function over the finite set of training data. For the sake of compactness of notation, however, we refrain from expressing the loss function as the sum of two terms, with the implicit understanding that Assumption~\ref{assumption finite value} is concerned only with the data-dependent component of $f(\cdot,\cdot)$. Finally, in contrast to Assumption~\ref{assumption strongly convex}, we make the following assumption in relation to analysis of the class of nonconvex loss functions.
\begin{assumptionprime}{assumption strongly convex}[Local strong convexity]\label{assumption positive definite hessian}
The loss function $f(\bw,\bz)$ is nonconvex and a.s.\ twice differentiable in the first argument. Next, let $\nabla^2 F(\bw)$ denote the Hessian of the statistical risk $F(\bw) := \E_{\bbP}[f(\bw,\bz)]$ and let $\cW_s^*$ denote the set of all first-order stationary points of $F(\bw)$, i.e., $\cW^*_s := \{\bw \in \R^d : \nabla F(\bw)=0\}$. Then, for any $\bw_s^* \in \cW_s^*$, the statistical risk is \emph{locally} $\lambda$-strongly convex in a sufficiently large neighborhood of $\bw_s^*$, i.e., there exist positive constants $\lambda$ and $\beta$ such that $\forall \bw \in \B(\bw^*_s,\beta), \nabla^2 F(\bw)\succeq\lambda \bI$.
\end{assumptionprime}
It is straightforward to see that the local strong convexity of the statistical risk does not imply the \emph{global} strong convexity of either the statistical risk or the loss function. \revise{Assumptions similar to Assumption~\ref{assumption positive definite hessian} are nowadays routinely used for analysis of nonconvex optimization problems in machine learning; see, e.g., \cite{Sun2015nonconvex,Jain2017nonconvex,Yonel2020,zhou2016noncovex}. In particular, Assumption~\ref{assumption positive definite hessian} along with the proof techniques utilized in this work allow the theoretical results to be applicable to a broader class of functions in which the strong convexity is preserved only locally.}

\revise{
\begin{remark}
The convergence guarantees for nonconvex loss functions in this paper are \emph{local} in the sense that they hold as long as the BRIDGE iterates are initialized within a sufficiently small neighborhood of a stationary point (cf.~Section~\ref{sec: theoretical analysis}). Such local convergence guarantees are typical of many results in nonconvex optimization (see, e.g.,~\cite{ypma1984local,ochs2018local,bock2019proof}), but they do not imply that an iterative algorithm requires knowledge of the stationary point.
\end{remark}
}

\subsection{System model for decentralized learning}
Consider a network of $M$ nodes (devices, machines, etc.), expressed as a directed, static, and connected graph $\cG(\cJ,\cE)$ in which the set $\cJ:=\{1,\dots,M\}$ represents nodes in the network and the set of edges $\cE$ represents communication links between the nodes. Specifically, $(j,i) \in \cE$ if and only if node $i$ can directly receive messages from node $j$ and vice versa. We also define the neighborhood $\cN_j$ of node $j$ as the set of nodes that can send messages to it, i.e., $\cN_j:=\lbrace i\in \cJ:(i,j)\in \cE\rbrace$. We assume each node $j$ has access to a local training dataset $\cZ_j := \lbrace \bz_{jn}\rbrace_{n=1}^{\vert \cZ_j\vert}$. For simplicity of exposition, we assume the cardinalities of the local training sets to be the same, as the generalization of our results to the case of $\cZ_j$'s not being same sized is trivial. Collectively, therefore, the network has a total of $MN$ samples that could be utilized for learning purposes.

In order to obtain an estimate of the statistical risk minimizer $\bw^*$ (cf.~\eqref{eqn: SRM.equation}) in this decentralized setting, one would ideally like to solve the following ERM problem:
\begin{align}\label{eqn: ERM}
\min\limits_{\bw\in \R^d} \frac{1}{MN}\sum\limits_{j=1}^M\sum\limits_{n=1}^N f(\bw,\bz_{jn}) = \min\limits_{\bw\in \R^d}\frac{1}{M}\sum\limits_{j=1}^M f_j(\bw),
\end{align}
where we have used $f_j(\bw) := \frac{1}{N} \sum_{n=1}^N f(\bw,\bz_{jn})$ to denote the \emph{local} empirical risk associated with the $j$-th node. In particular, it is well known that the minimizer of \eqref{eqn: ERM} will statistically converge with high probability to $\bw^*$ in the case of a strictly convex loss function~\cite{Vapnik2013nature}. The problem in \eqref{eqn: ERM}, however, necessitates bringing together of the data at a single location; as such, it cannot be practically solved in its current form in the decentralized setting. Instead, we assume each node $j$ maintains a local version $\bw_j$ of the desired global model and collaborate among themselves to solve the following \emph{decentralized} ERM problem:
\begin{align}\label{eqn: decentralized ERM}
    \min\limits_{\{\bw_1,\dots,\bw_M\}} \frac{1}{M}\sum\limits_{j=1}^M f_j(\bw_j) \ \text{subject to} \ \forall i,j, \ \bw_i = \bw_j.
\end{align}
Traditional decentralized learning algorithms proceed iteratively to solve this decentralized ERM problem \cite{predd2006distributed,forero2010consensus,boyd2011distributed,duchi2012dual,AliH.Sayed2014,nedic2018,sun2021decentralized}. This is typically accomplished through each node $j$ engaging in two tasks during each iteration: update the local variable $\bw_j$ according to some (local and data-dependent) rule $g_j(\cdot)$, and broadcast some summary of its local information to the nodes that have node $j$ in their respective neighborhoods.

\subsection{Byzantine-resilient decentralized learning}
While decentralized learning is well understood in the case of faultless networks, the main assumption in this paper is that some of the network nodes can arbitrarily deviate from their intended behavior during the iterative process. Such deviations could be caused by malfunctioning equipment, cyberattacks, etc. We model the deviations of the faulty nodes as a \emph{Byzantine failure}, which is formally defined as follows~\cite{Lamport1982byzantine,Driscoll2003byzantine}.
\begin{definition}[Byzantine node]
A node $j \in \cJ$ is said to have undergone a Byzantine failure if, during any iteration of decentralized learning, it either updates its local variable $\bw_j$ using an update rule $\wtg_j(\cdot)\neq g_j(\cdot)$ or it broadcasts some information other than the intended summary of its local information to the nodes in its vicinity.
\end{definition}

Throughout the remainder of this paper, we use $\cR \subseteq \cJ$ and $\cB \subset \cJ$ to denote the sets of nonfaulty and Byzantine nodes in the network, respectively. In addition, we use $r$ to denote the cardinality of the set $\cR$ and assume that the number of Byzantine nodes is upper bounded by an integer $b$. Thus, we have $0 \leq |\cB| \leq b$ and $r \geq M - b$. In addition, without loss of generality, we label the nonfaulty nodes from $1$ to $r$ within our analysis, i.e., $\cR := \{1,\dots,r\}$.

Under this assumption of Byzantine failures in the network, it is straightforward to see that the decentralized ERM problem as stated in \eqref{eqn: decentralized ERM} cannot be solved. Rather, the best one could hope for is to solve an ERM problem that is restricted to the set of nonfaulty nodes, i.e.,
\begin{align}\label{eqn: restricted decentralized ERM}
    \min\limits_{\{\bw_j : j \in \cR\}} \frac{1}{r}\sum\limits_{j \in \cR} f_j(\bw_j) \ \text{subject to} \ \forall i,j \in \cR, \ \bw_i = \bw_j,
\end{align}
except that the set $\cR$ is unknown to an algorithm and therefore traditional decentralized learning algorithms cannot be utilized for this purpose. Consequently, the main goal in this paper is threefold: ($i$) Develop a decentralized learning algorithm that can provably solve some variant of the decentralized ERM problem \eqref{eqn: decentralized ERM}; ($ii$) Establish that the resulting solution statistically converges to the statistical risk minimizer (Assumption~\ref{assumption strongly convex}) or a stationary point of the statistical risk (Assumption~\ref{assumption positive definite hessian}); and \revise{($iii$) Characterize the sample complexity of the solution, i.e., the statistical rate of convergence as a function of the number of samples, $rN$, associated with the nonfaulty nodes}.

In order to accomplish the stated goal of this paper, we need to make one additional assumption concerning the topology of the network. This assumption, which is common in the literature on Byzantine resilience within decentralized networks~\cite{Su2016fault,yang2019byrdie}, requires definitions of the notions of a \emph{source component} of a graph and a \emph{reduced graph}, $\cG_{\textsf{red}}(b)$, of $\cG$.
\begin{definition}[Source component]
A source component of a graph is any subset of graph nodes such that each node in the subset has a directed path to every other node in the graph.
\end{definition}
\begin{definition}[Reduced graph]
A subgraph $\cG_{\textsf{red}}(b)$ of $\cG$ is called a reduced graph with parameter $b$ if it is generated from $\cG$ by ($i$) removing all Byzantine nodes along with all their incoming and outgoing edges from $\cG$, and ($ii$) additionally removing $b$ incoming edges from each nonfaulty node.
\end{definition}
\begin{assumption}[Sufficient network connectivity]\label{assumption reduced graph}
The decentralized network is assumed to be sufficiently connected in the sense that all reduced graphs $\cG_\textsf{red}(b)$ of the underlying graph $\cG(\cJ,\cE)$ contain at least one source component of cardinality greater than or equal to $(b+1)$.
\end{assumption}

We conclude by expanding further on Assumption~\ref{assumption reduced graph}, which concerns the redundancy of information flow within the network. In words, this assumption ensures that each nonfaulty node can continue to receive information from a few other nonfaulty nodes even after a certain number of edges have been removed from every nonfaulty node. And while efficient certification of this assumption remains an open problem, there is an understanding of the generation of graphs that satisfy this assumption~\cite{leblanc2013resilient}. In addition, we have empirically observed that Assumption~\ref{assumption reduced graph} is often satisfied in Erd\H{o}s--R\'{e}nyi graphs as long as the degree of the least connected node is larger than $2b$. This is also the approach we take while generating graphs for our numerical experiments.

\revise{
\begin{remark}
In the finite sample regime, in which each (nonfaulty) node has only a finite number of training data samples, the local empirical risk $f_j(\bw)$ at every node will be different due to the randomness of the data samples, regardless of whether the training data across the nodes are i.i.d.\ or non-i.i.d. While this makes the formulation in \eqref{eqn: restricted decentralized ERM} similar to the one in \cite{sundaram2018distributed} and \cite{Kuwaranancharoen2020ByzantineResilientDO} for scalar-valued and vector-valued Byzantine-resilient decentralized optimization, respectively, the fundamental difference between the statistical learning framework of this work and the optimization-only framework in \cite{sundaram2018distributed,Kuwaranancharoen2020ByzantineResilientDO} is the intrinsic focus on sample complexity in statistical learning. In particular, the sample complexity results in Section~\ref{ssec:optimality.analysis} also help characterize the gap between \emph{local-only learning}, in which every regular node learns its own model using its own training data, and decentralized learning, in which the regular nodes collaborate to learn a common model.
\end{remark}
}

\section{Byzantine-resilient Decentralized\\Gradient Descent}\label{section BRIDGE}
In the faultless case, the decentralized ERM problem \eqref{eqn: decentralized ERM} can be solved, perhaps to one of its stationary points, using any one of the distributed/decentralized optimization methods in the literature~\cite{nedic2009,ram2010distributed,nedic2015distributed,nedic2020,Lorenzo2016NEXTIN,Zeng2018OnND,Sun2019ImprovingTS,xin2021fast}. The prototypical \emph{distributed gradient descent} (DGD) method~\cite{nedic2009} with decreasing step size, for instance, accomplishes this for (strongly) convex loss functions by letting each node in iteration $(t+1)$ update its local variable $\bw_j(t)$ as
\begin{align}\label{eqn: dgd}
\bw_j(t+1)=\sum\limits_{i\in \cN_j\cup \lbrace j\rbrace}a_{ji}\bw_i(t)-\rho(t)\nabla f_j(\bw_j(t)),
\end{align}
where $0 \leq a_{ji} \leq 1$ is the weighting that node $j$ applies to the local variable $\bw_i(t)$ that it receives from node $i$, and $\{\rho(t)\}$ denotes a positive sequence of step sizes that satisfies $\rho(t+1)\leq \rho(t)$, $\rho(t) \stackrel{t}{\rightarrow} 0$, $\sum_{t=0}^\infty \rho(t) = \infty$, and $\sum_{t=0}^\infty \rho^2(t)<\infty$. One choice for such a sequence is $\rho(t)=\frac{1}{\lambda(t_0+t)}$ for some $t_0$, which ensures that a network-wide consensus is reached among all nodes, i.e., $\forall i,j, \bw_i(t) \stackrel{t}{\to} \bw_j(t)$, and all local variables converge to the decentralized (and thus the centralized) ERM solution.

Traditional distributed/decentralized optimization methods, however, fail to reach a stationary point of the decentralized ERM problem \eqref{eqn: decentralized ERM} (or its restricted variant \eqref{eqn: restricted decentralized ERM}) in the presence of a single Byzantine failure in the network~\cite{su2015byzantine,sundaram2018distributed,mhamdi2018hidden,yin2018byzantine,alistarh2018byzantine,Yang2019ByzantineResilientSG,Data2020ByzantineResilientSI}. To overcome this shortcoming of the traditional approaches as well as improve on the limitations of existing works on Bynzatine-resilient decentralized learning (cf.~Sections~\ref{ssec:prior.work} and \ref{ssec:our.contributions}), we introduce an algorithmic framework termed \textbf{B}yzantine-\textbf{r}es\textbf{i}lient \textbf{d}ecentralized \textbf{g}radient d\textbf{e}scent (BRIDGE).

The BRIDGE framework, which is listed in Algorithm~\ref{gradient descent algorithm}, is a gradient descent-based approach whose main update step (Step~\ref{alg:BRIDGE.update} in Algorithm~\ref{gradient descent algorithm}) is similar to the DGD update \eqref{eqn: dgd}. The main difference between the BRIDGE framework and DGD is that each node $j \in \cR$ in BRIDGE \emph{screens} the incoming messages from its neighboring nodes for \emph{potentially} malicious content (Step~\ref{alg:BRIDGE.screen} in Algorithm~\ref{gradient descent algorithm}) before updating its local variable $\bw_j(t)$. Note, however, that BRIDGE does not permanently label any nodes as malicious, which also allows it to manage any transitory Byzantine failures in a graceful manner. While this makes BRIDGE similar to the ByRDiE algorithm~\cite{yang2019byrdie}, the fundamental advantage of BRIDGE over ByRDiE is its scalability that comes from the fact that it eschews the one-coordinate-at-a-time update of the local variables in ByRDiE in favor of one update of the entire vector $\bw_j(t)$ in each iteration. In terms of other details, the BRIDGE framework is input with the maximum number of Byzantine nodes $b$ that need to be tolerated, a decreasing step size sequence $\{\rho(t)\}$, and the maximum number of gradient descent iterations $t_{\max}$. Next, the local variable at each nonfaulty node $j$ in the network is initialized at $\bw_j(0)$. Afterward, within every iteration $(t+1)$ of the framework, each node $j \in \cR$ broadcasts $\bw_j(t)$ as well as receives $\bw_i(t)$ from every node $i \in \cN_j$. This is followed by every node $j \in \cR$ screening the received $\bw_i(t)$'s for any malicious information and then updating the local variable $\bw_j(t)$.

\begin{algorithm}[t]
\caption{The BRIDGE Framework} \label{gradient descent algorithm}
\begin{algorithmic}[1]
\REQUIRE Local datasets $\cZ_j$, maximum number of Byzantine nodes $b$, step size sequence $\lbrace\rho(t)\rbrace_{t=0}^{\infty}$, and maximum number of iterations $t_{\max}$
\STATE \textbf{Initialize:} $t \leftarrow 0$ and $\forall j \in \cR, \bw_j(0)$
\FOR{$t=0, 1, \dots, t_{\max}-1$}
\STATE Broadcast $\bw_j(t), \forall j\in \cR$
\STATE Receive $\bw_i(t)$ at each node $j \in \cR$ from \\ \qquad every $i\in \cN_j \subset (\cR \cup \cB)$
\STATE $\by_j(t)\leftarrow \textsf{screen}(\{\bw_i(t)\}_{i\in \cN_j\cup \{j\}}), \forall j \in \cR$ \label{alg:BRIDGE.screen}
\STATE $\bw_j(t+1)\leftarrow\by_j(t)-\rho(t)\triangledown f_j(\bw_j(t)), \forall j \in \cR$ \label{alg:BRIDGE.update}
\ENDFOR\label{for loop end}
\ENSURE $\bw_j(t_{\max}), \forall j \in \cR$
\end{algorithmic}
\end{algorithm}

\begin{table*}[t]
\centering
\begin{tabular}{|c c c c|}
 \hline
 \bf Variant & \bf Screening & \bf Min.\ neighborhood size & \bf Ave.\ computational complexity\\
 \hline\hline
 BRIDGE-T~  & coordinate-wise & $2b+1$ & $\cO(nd)$, where  $n := \max_j |\cN_j|$\\
 BRIDGE-M  & coordinate-wise &  $1$ & $\cO(nd)$\\
 BRIDGE-K & vector & $b+3$ & $\cO(n^2d)$  \\
 BRIDGE-B & vector + coordinate-wise & $\max(4b,3b+2)+1$ & $\cO(n^2d)$  \\
 \hline
\end{tabular}
\caption{Comparison between the four different variants of the BRIDGE framework.}
\label{table:12}
\end{table*}

In the following, we discuss four different variants of the \emph{screening rule} (Step~\ref{alg:BRIDGE.screen} in Algorithm~\ref{gradient descent algorithm}), each one of which in turn gives rise to a different realization of the BRIDGE framework. The motivation for these screening rules comes from the literature on robust statistics~\cite{huber2011robust}, with all these rules appearing in some form within the literature on robust averaging consensus~\cite{leblanc2013resilient,vaidya2014iterative} and robust distributed learning ~\cite{blanchard2017machine,yin2018byzantine,alistarh2018byzantine,mhamdi2018hidden}. The challenge here of course, as discussed in Section~\ref{section introduction}, is that these prior works on robust averaging consensus and distributed learning do not translate into equivalent results for Byzantine-resilient decentralized learning.

The \textbf{BRIDGE-T} variant of the BRIDGE framework uses the coordinate-wise trimmed mean as the screening rule. A similar screening principle has been utilized within distributed frameworks~\cite{yin2018byzantine} and decentralized frameworks~\cite{Su2016fault,sundaram2018distributed,yang2019byrdie}. The coordinate-wise trimmed-mean screening within BRIDGE-T filters the $b$ largest and the $b$ smallest values in each coordinate of the local variables $\bw_i(t)$ received from the neighborhood of node $j$ and uses an average of the remaining values for the update of $\bw_j(t)$. Specifically, for any iteration index $t$, BRIDGE-T finds the following three sets for each coordinate $k \in \{1,\dots,d\}$ in parallel:
\begin{align}
\overline{\cN}_j^{k}(t) &:=\argmin\limits_{\cX: \cX\subset \cN_j, \vert \cX\vert =b }\sum\limits_{i\in \cX}[\bw_i(t)]_k,\\
\underline{\cN}_j^{k}(t) &:=\argmax\limits_{\cX: \cX\subset \cN_j, \vert \cX\vert =b }\sum\limits_{i\in \cX}[\bw_i(t)]_k, \quad \text{and}\\
\revise{\cC_j^k(t)} &:= \cN_j \setminus\left\{\overline{\cN}_j^{k}(t)\bigcup\underline{\cN}_j^{k}(t)\right\}.
\end{align}
Afterward, the screening routine outputs a combined and filtered vector $\by_j(t)$ whose $k$-th element is given by
\begin{align}\label{eqn:BRIDGE.T.filtered.y}
[\by_j(t)]_k = \frac{1}{\vert\cN_j\vert -2b+1}\sum\limits_{i\in \revise{\cC_j^k(t)}\cup  \{j\}}[\bw_i(t)]_k.
\end{align}
Notice that BRIDGE-T requires each node to have at least $2b + 1$ neighbors. Also, note that the elements from different neighbors may survive the screening at different coordinates. Therefore, the average within BRIDGE-T is not taken over vectors; rather, the calculation of $\by_j(t)$ has to be carried out in a coordinate-wise manner.

The \textbf{BRIDGE-M} variant uses the coordinate-wise median as the screening rule, with a similar screening idea having been utilized within distributed frameworks~\cite{yin2018byzantine}. Similar to BRIDGE-T, BRIDGE-M is also a coordinate-wise screening procedure in which the $k$-th element of the combined and filtered output $\by_j(t)$ takes the form
\begin{align}
[\by_j(t)]_k = \textsf{median}\bigg(\{[\bw_i(t)]_k\}_{i\in \cN_j\cup  \{j\}}\bigg).
\end{align}
Notice that, unlike BRIDGE-T, the coordinate-wise median screening within BRIDGE-M neither requires an explicit knowledge of $b$ nor does it impose an explicit constraint on the minimum number of neighbors of each node.

The \textbf{BRIDGE-K} variant uses the \emph{Krum function} as the screening rule, which is similar to the screening principle that has been employed within distributed frameworks~\cite{blanchard2017machine}. In terms of specifics, the Krum screening for the decentralized framework can be described as follows. Given $i, h \in \cN_j\cup\{j\}$, write $h \sim i$ if $\bw_h(t)$ is one of the $|\cN_j| - b - 2$ vectors with the smallest Euclidean distance, expressed as $\|\bw_h(t) - \bw_i(t)\|$, from $\bw_i(t)$. The Krum-based screening at node $j$ then finds the neighbor index $i_j^*(t)$ as
\begin{align}\label{eqn: Krum}
i_j^*(t) = \argmin\limits_{i\in \cN_j}\sum\limits_{h \in \cN_j\cup\{j\}:h\sim i} \|\bw_h(t) - \bw_i(t)\|,
\end{align}
and outputs the (combined and filtered) vector $\by_j(t)$ as $\by_j(t)=\bw_{i_j^*}(t)$. Unlike BRIDGE-T and BRIDGE-M, BRIDGE-K utilizes vector-valued operations for screening, resulting in the surviving vector $\by_j(t)$ to be entirely from one neighbor of each node. The Krum screening rule within BRIDGE-K requires the neighborhood of every node to be larger than $b+2$. Note that since the Krum function requires the pairwise distances of all nodes within the neighborhood of every node, BRIDGE-K has high computational complexity in comparison to BRIDGE-T and BRIDGE-M.

Last but not least, the \textbf{BRIDGE-B} variant---inspired by a similar screening procedure within distributed frameworks~\cite{mhamdi2018hidden}---uses a combination of Krum and coordinate-wise trimmed mean as the screening rule. Specifically, the screening within BRIDGE-B involves first selecting $|\cN_j| - 2b$ neighbors of the $j$-th node by recursively finding an index $i_j^*(t) \in \cN_j$ using \eqref{eqn: Krum}, removing the selected node from the neighborhood, finding a new index from $\cN_j \setminus \{i_j^*(t)\}$ using \eqref{eqn: Krum} again, and repeating this Krum-based process $|\cN_j| - 2b$ times. Next, coordinate-wise trimmed mean-based screening, as described within BRIDGE-T, is applied to the received $\bw_i(t)$'s of the $|\cN_j| - 2b$ neighbors of node $j$ that survive the first-stage Krum-based screening. Intuitively, the Krum-based vector screening first guarantees the surviving neighbors have the closest $\bw_i(t)$'s in terms of the Euclidean distance, while coordinate-wise trimmed-mean screening afterward guarantees that each coordinate of the combined and filtered vector $\by_j(t)$ only includes the ``inlier'' values. The cost of this two-step screening procedure includes high computational complexity due to the use of the Krum function and the stricter requirement that the neighborhood of each node be larger than $\max(4b,3b+2)$.

We conclude by providing a comparison in Table~\ref{table:12} between the four different variants of the BRIDGE framework. Note that both BRIDGE-T and BRIDGE-B reduce to DGD in the case of $b=0$; this, however, is not the case for BRIDGE-M and BRIDGE-K. It is also worth noting that additional variants of BRIDGE can be obtained through further combinations of the different screening rules and/or incorporation of additional ideas from the literature on robust statistics. Nonetheless, each variant of the BRIDGE framework requires its own theoretical analysis for convergence guarantees. In this paper, we limit ourselves to BRIDGE-T for this purpose and provide the corresponding guarantees in the next section.

\section{BRIDGE-T: Convergence Guarantees}\label{sec: theoretical analysis}
In this section, we derive the algorithmic and statistical convergence guarantees for BRIDGE-T for both convex and nonconvex loss functions. \revise{While these results match those for ByRDiE for convex loss functions, they cannot be obtained directly from~\cite{yang2019byrdie} since the filtered vector $\by_j(t)$ in BRIDGE-T corresponds to an iteration-dependent amalgamation of the iterates $\{\bw_i(t)\}$ in the neighborhood of node $j$ (cf.~\eqref{eqn:BRIDGE.T.filtered.y}).} Our statistical convergence guarantees require the training data to be independent and identically distributed (i.i.d.) among all the nodes. Additionally, let $\bw^*$ denote the unique global minimizer of the statistical risk in the case of the strongly convex loss function, while it denotes one of the first-order stationary points of the statistical risk in the case of the nonconvex loss function. We assume there exists a positive constant $\Gamma$ such that for all $j \in \cR,t \in \R^d$, $\|\bw_j(t)-\bw^*\|\leq \Gamma$. Note that this $\Gamma$ can be arbitrary large and so the above assumption, which is again needed for derivation of the statistical rate of convergence, is a mild one (also, see Lemma~\ref{lemma:condition for not going out} and its accompanying discussion). Broadly speaking, the analysis for our algorithmic and statistical convergence guarantees proceeds as follows.

\looseness=-1 First, we define a ``consensus'' vector $\bv(t)$ and establish in Section~\ref{ssec:consensus.analysis} that $\forall j \in \cR, \bw_j(t)\rightarrow\bv(t)$ as $t\rightarrow\infty$ for an appropriate choice of the step-size sequence $\rho(t)$ that satisfies $\rho(t+1)\leq \rho(t)$, $\rho(t) \stackrel{t}{\rightarrow} 0$, $\sum_{t=0}^\infty \rho(t) = \infty$, and $\sum_{t=0}^\infty \rho^2(t)<\infty$. In particular, we work with the step size $\rho(t)=\frac{1}{\lambda(t_0+t)}, t_0 \geq \frac{L}{\lambda}$, where $L$ is the Lipschitz constant of the loss function; see, e.g.,  Assumption~\ref{assumption lipschitz}. This consensus analysis relies on the condition that the $\bw_j(0)$'s for all $j\in \cR$ are initialized such that $\bv(0) \in \B(\bw^*,\Gamma)$. One such choice of initialization is initializing all $\bw_j(0)$'s to be within $\B(\bw^*,\Gamma)$. (Note that in terms of our analysis for the nonconvex loss function, which is provided in Section~\ref{nonconvex analysis}, we will impose additional constraints on the initialization.). Afterward, we define two vectors $\bu(t+1)$ and $\bx(t+1)$ in Section~\ref{ssec:optimality.analysis} that correspond to gradient descent updates of the consensus vector $\bv(t)$ using, respectively, a convex combination of gradients of the local loss functions evaluated at $\bv(t)$ and the gradient of the statistical risk evaluated at $\bv(t)$. Finally, we define the following collection of distances: $a_1(t) := \|\bx(t+1)-\bw^*\|$, $a_2(t) := \|\bu(t+1)-\bx(t+1)\|$, $a_3(t) := \|\bv(t+1)-\bu(t+1)\|$, and $a_4(t+1) := \max_{j\in\cR}\|\bw_j(t+1)-\bv(t+1)\|$. It is straightforward to see that $\|\bv(t+1)-\bw^*\| \leq a_1(t)+a_2(t)+a_3(t)$, and we establish in Section~\ref{ssec:optimality.analysis} that $a_1(t)+a_2(t)+a_3(t) \stackrel{t}{\to} 0$ with high probability as well as $a_4(t+1)\stackrel{t}{\to} 0$ using Assumptions \ref{assumption lipschitz}, \ref{assumption finite value}, \ref{assumption strongly convex} and \ref{assumption reduced graph} in the convex setting and Assumptions \ref{assumption lipschitz}, \ref{assumption finite value},  \ref{assumption positive definite hessian} and \ref{assumption reduced graph} in the nonconvex setting, thereby completing our proof of optimality for both convex and nonconvex loss functions.

\subsection{Consensus guarantees}\label{ssec:consensus.analysis}
Let us pick an arbitrary index $k \in \{1,\dots,d\}$ and define a vector $\bOmega(t)\in \R^{r}$ whose respective elements correspond to the $k$-th element of the iterate $\bw_j(t)$ of nonfaulty nodes, i.e., $\forall j \in \cR, [\bOmega(t)]_j=[\bw_j(t)]_k$. Note that $\bOmega(t)$ as well as most of the variables in our discussion in this section depend on the index $k$; however, since $k$ is arbitrary, we drop this explicit dependence on $k$ in many instances for simplicity of notation.

We first show that the BRIDGE-T update at the nonfaulty nodes in the $k$-th coordinate can be expressed in a form that only involves the nonfaulty nodes. Specifically, we write
\begin{align}\label{eqn: nonfaulty.update}
    \bOmega(t+1)=\bY(t)\bOmega(t)-\rho(t)\bg(t),
\end{align}
where the vector $\bg(t)$ is defined as $[\bg(t)]_j=[\nabla f_j(\bw_j(t))]_k, j \in \cR$. The formulation of the matrix $\bY(t)$ in this expression is as follows. Let $\cN_j^r$ denote the nonfaulty nodes in the neighborhood of node $j$, i.e., $\cN_j^r := \cR\cap \cN_j$. The set of Byzantine neighbors of node $j$ can then be defined as $\cN_j^b := \cN_j\setminus\cN_j^r$. To make the rest of the expressions clearer, we drop the iteration index $t$ for the remainder of this discussion, even though the variables are still $t$-dependent. \revise{Let us now define the notation $b^* := |\cB|$ as the actual (unknown) number of Byzantine nodes in the network, $b_j^k$ as the number of Byzantine nodes remaining in the filtered set  $\cC_j^k$, and $q_j^k := b - b^* + b_j^k$. Since $b - b^* \geq 0$ by assumption and $b_j^k \geq 0$ by definition, notice that only one of two cases can happen during each iteration for every coordinate $k$: ($i$) $q_j^k > 0$ or ($ii$) $q_j^k = 0$. For case ($i$), we either have $b - b^* > 0$ or $b_j^k > 0$ or both. These conditions correspond to the scenario where the node $j$ filters out more than $b$ regular nodes from its neighborhood.} Thus, we know that $\overline{\cN}_j^{k}\cap\cN_j^r\neq\emptyset$. Likewise, it follows that $\underline{\cN}_j^{k}\cap\cN_j^r\neq\emptyset$. Then $\exists m_j'\in \overline{\cN}_j^{k}\cap\cN_j^r$ and $m_j''\in \underline{\cN}_j^{k}\cap\cN_j^r$ satisfying $[\bw_{m_j'}]_k\leq[\bw_i]_k\leq[\bw_{m_j''}]_k$ for any $i\in \revise{\cC_j^k}$. Thus, for every $i\in \revise{\cC_j^k}\cap\cN_j^b$, $\exists \theta_i\in (0,1)$ satisfying $[\bw_i]_k=\theta_i [\bw_{m_j'}]_k+(1-\theta_i)[\bw_{m_j''}]_k$. Consequently, the elements of the matrix $\bY$ can be written as
\begin{equation}\label{elements in M}
[\bY]_{ji}=\begin{cases}
\frac{1}{\revise{2}(\vert\cN_j\vert-2b+1)},& i\in\cN^r_j\cap\revise{\cC_j^k},\\
\frac{1}{\vert\cN_j\vert-2b+1},&i=j,\\
\sum\limits_{i'\in\cN^b_j\cap\revise{\cC_j^k}}\frac{\theta_{i'}}{\revise{q_j^k}(\vert\cN_j\vert-2b+1)}\\\revise{\qquad +\sum\limits_{i'\in\cN^r_j\cap\cC_j\revise{^k}}\frac{\theta_{i'}}{\revise{q_j^k}(\vert\cN_j\vert-2b+1)}},&i\in \overline{\cN}_j^{k}\cap\cN_j^r,\\
\sum\limits_{i'\in\cN^b_j\cap\revise{\cC_j^k}}\frac{1-\theta_{i'}}{\revise{q_j^k}(\vert\cN_j\vert-2b+1)}\\\revise{\qquad+\sum\limits_{i'\in\cN^r_j\cap\cC_j\revise{^k}}\frac{1-\theta_{i'}}{\revise{q_j^k}(\vert\cN_j\vert-2b+1)}},&i\in \underline{\cN}_j^{k}\cap\cN_j^r,\\
0,&\text{otherwise}.
\end{cases}
\end{equation}
\revise{For case ($ii$), we must have that $b-b^* = 0$ and $b_j^k = 0$. Thus, all the filtered nodes in $\cC_j^k$ would be regular nodes in this case.} Therefore, we can describe $\bY$ in this case \revise{as
\begin{equation}\label{elements in MM}
[\bY]_{ji}=\begin{cases}
\frac{1}{\vert\cN_j\vert-2b+1},&i\in \{j\}\cup \cC_j^k,\\
0,&\text{otherwise}.
\end{cases}
\end{equation}
Combining the expressions of $\bY$ in the two cases above} allows us to express the update in \eqref{eqn: nonfaulty.update} exclusively in terms of information from the nonfaulty nodes.

Next, we define $\psi$ to be the total number of reduced graphs that can be generated from $\cG$, the parameter $\nu$ as $\nu:=\psi r$, and the maximum neighborhood size of the nonfaulty nodes as $\cN_{\max} := \max_{j\in \cR}\vert\cN_j\vert$. Further, we define a transition matrix $\bPhi(t,t_0)$ from some index $t_0 \leq t$ to $t$, i.e.,
\begin{equation}
    \bPhi(t,t_0) :=\bY(t)\bY(t-1)\cdots\bY(t_0).
\end{equation}
Then it follows from \cite[Lemma 4]{Su2015fault} that if Assumption~\ref{assumption reduced graph} is satisfied then
\begin{align}
\lim\limits_{t\rightarrow\infty}\bPhi(t,t_0)=\bone{\balpha}^T(t_0),
\end{align}
where the vector ${\balpha}(t_0) \in \R^r$ satisfies $[{\balpha}(t_0)]_j\geq 0$ and $\sum_{j=1}^{r}[{\balpha}(t_0)]_j=1$. In particular, we have \cite[Theorem 3]{Su2015fault}
\begin{equation} \label{13}
\left|[\bPhi(t,t_0)]_{ji}-[{\balpha}(t_0)]_i\right|\leq \mu^{(\frac{t-t_0+1}{\nu})},
\end{equation}
where $\mu \in (0,1)$ is defined as $\mu := 1-\frac{1}{(2\cN_{\max}-2b+1)^\nu}$.

Next, it follows from \eqref{eqn: nonfaulty.update} and the definition of $\bPhi(t,t_0)$ that
\begin{align}
\bOmega(t)&=\bY(t-1)\bOmega(t-1)-\rho(t-1)\bg(t-1)\nonumber\\
&=\bY(t-1)\bY(t-2)\cdot\cdot\cdot \bY(0)\bOmega(0)\nonumber\\&\qquad-\sum\limits_{\tau=0}^{t-1} \bY(t-1)\bY(t-2)\cdots \bY(\tau+1)\rho(\tau)\bg(\tau)\nonumber
\\&=\bPhi(t-1,0)\bOmega(0)-\sum\limits_{\tau=0}^{t-1}\bPhi(t-1,\tau+1)\rho(\tau)\bg(\tau).
\end{align}
Now, similar to \cite[Convergence Analysis of Algorithm 1]{Su2015fault}, suppose all nodes stop computing their local gradients after iteration $t$ so that $\bg(\tau)=0$ when $\tau>t$. Note that this is without loss of generality when we let $t$ approach infinity, as we recover BRIDGE-T in that case. Further, let $T \geq 0$ be an integer and define a vector $\bar{\bv}(t)$ as follows:
\begin{align}\label{define vt}
&\bar{\bv}(t)=\lim\limits_{T\to\infty}\bOmega(t+T+1)\nonumber\\
&=\lim\limits_{T\to\infty}\bPhi(t+T,0)\bOmega(0)-\lim\limits_{T\to\infty}\sum\limits_{\tau=0}^{t+T}\bPhi(t+T,\tau\revise{+1})\rho(\tau)\bg(\tau)\nonumber\\
&=\bone{\balpha}^T(0)\bOmega(0)-\sum\limits_{\tau=0}^{t-1}\bone{\balpha}^T(\tau\revise{+1})\rho(\tau)\bg(\tau).
\end{align}
Notice that $\bar{\bv}(t)$ is a constant vector and we define a scalar-valued sequence $v(t)$ to be any one of its elements. We now show that $[\bw_j(t)]_k \stackrel{t}{\to} v(t)$. Indeed, we have from \eqref{define vt} that
\begin{align}\label{eqn: consensu.vector.entry}
v(t)&=\sum\limits_{i=1}^{r}[{\balpha}(0)]_i[\bw_i(0)]_k\nonumber\\&\qquad-\sum\limits_{\tau=0}^{t-1}\rho(\tau)\sum\limits_{i=1}^{r}[{\balpha}(\tau\revise{+1})]_i[\nabla f_i(\bw_i(\tau))]_k.
\end{align}
Also recall from the update of $[\bw_j(t)]_k$ that
\begin{align*}
[\bw_j(t)]_k&=\sum\limits_{i=1}^{r}[\bPhi(t-1,0)]_{ji}[\bw_i(0)]_k\nonumber\\&\qquad-\sum\limits_{\tau=0}^{t-1}\rho(\tau)\sum\limits_{i=1}^{r}[\bPhi(t-1,\tau\revise{+1})]_{ji}[\nabla f_i(\bw_i(\tau))]_k.
\end{align*}

From Assumption~\ref{assumption lipschitz} and the initialization of $\bw_j(t)$'s,  there exist two scalars $C_w$ and $L$ such that $\forall j\in \cR$, $\left\vert [\bw_j(0)]_k\right\vert\leq C_w$ and $\left\vert [\nabla f_j(\bw_j)]_k\right\vert\leq L$. Therefore, we have
\begin{align}\label{eqn: consensus in one dim}
&\vert [\bw_j(t)]_k-v(t)\vert \leq\vert\sum\limits_{i=1}^{r}([\bPhi(t-1,0)]_{ji}-[{\balpha}(0)]_i)[\bw_i(0)]_k\vert\nonumber \\&\quad+ \vert\sum\limits_{\tau=0}^{t-1}\rho(\tau)\sum\limits_{i=1}^{r}([\bPhi(t-1,\tau\revise{+1})]_{ji}-[{\balpha}(\tau\revise{+1})]_i)[\nabla f_i(\bw_i(\tau))]_k\vert\nonumber
\\&\leq r C_w\mu^{\frac{t}{\nu}}+r L\sum\limits_{\tau=0}^{t}\rho(\tau)\mu^{\frac{t-\tau+1}{\nu}} \stackrel{t}{\to} 0.
\end{align}
Here, the fact that the second term in the second inequality of (\ref{eqn: consensus in one dim}) converges to zero follows from our assumptions on the decreasing step size sequence along with \cite[Lemma~6]{Su2015fault}.

Finally, recall that the vector-valued iterate updates in BRIDGE-T can be thought of as individual updates of the $d$ coordinates in parallel. Therefore, since the coordinate $k$ was arbitrarily picked, we have proven that BRIDGE-T achieves consensus among the nonfaulty nodes for both convex and nonconvex loss functions, as summarized in the following.
\begin{theorem}\label{theorem: consensus a_4 distance}
Define a vector $\bv(t) \in \R^d$ as one whose $k$-th entry $[\bv(t)]_k$ is given by the right-hand-side of \eqref{eqn: consensu.vector.entry}. If Assumptions \ref{assumption lipschitz}  and \ref{assumption reduced graph} are satisfied, then the gap between $\bw_j(t), \forall j \in \cR,$ and $\bv(t)$ goes to $0$ as $t\to\infty$, i.e.,
\begin{align}\label{eqn: w converges to v}
&\lim_{t\to\infty}a_4(t) = \lim_{t\to\infty}\max_{j \in \cR}\| \bw_j(t)-\bv(t)\| \nonumber \\ &\quad \leq \lim_{t\to\infty}\left[\sqrt{d}r C_w\mu^{\frac{t}{\nu}}+\sqrt{d}r L\sum\limits_{\tau=0}^{t}\rho(\tau)\mu^{\frac{t-\tau+1}{\nu}}\right]= 0.
\end{align}
\end{theorem}
We conclude our discussion with a couple of remarks. First, note that Theorem~\ref{theorem: consensus a_4 distance} has been obtained without needing Assumptions~\ref{assumption finite value} and \ref{assumption strongly convex} / \ref{assumption positive definite hessian}. Thus, BRIDGE-T guarantees consensus among the nonfaulty nodes for both convex and nonconvex loss functions under a general set of assumptions. Second, notice that the second term in \eqref{eqn: w converges to v} is a sum of $t$ terms and it converges to $0$ at a slower rate than the first term. Among all the sub-terms in the sum of the second term, the last term $\rho(t)\mu^\frac{1}{\nu}$ is the one that converge to zero at the slowest rate. Thus, the rate at which BRIDGE-T achieves consensus is determined by this sub-term and is given by $\cO(\sqrt{d}\rho(t))$. In particular, if we choose $\rho(t)$ to be $\cO(1/t)$, the rate of consensus for BRIDGE-T is $\cO(\sqrt{d}/t)$.

\subsection{Statistical optimality guarantees}\label{ssec:optimality.analysis}
While Theorem~\ref{theorem: consensus a_4 distance} guarantees consensus among the nonfaulty nodes by providing an upper bound on the distance $a_4(t)$, this result alone cannot be used to characterize the gap between the iterates $\{\bw_j(t)\}_{j\in\cR}$ and the global minimizer (resp., first-order stationary point) $\bw^*$ of the statistical risk for the convex loss function (resp., nonconvex loss function). We accomplish the goal of establishing the statistical optimality by providing bounds on the remaining three distances $a_1(t)$, $a_2(t)$, and $a_3(t)$ described in the beginning of the section.

We start with a bound on $a_3(t)$. To this end, let $\bv(t)$ be as defined in Theorem~\ref{theorem: consensus a_4 distance} and notice from \eqref{define vt} that
\begin{align}
    \bv(t+1)=\bv(t)-\rho(t)\bg_1(t),
\end{align}
where $\bg_1(t)$ has the $k$-th entry defined in terms of gradients of the local loss functions evaluated at the local iterates as $[\bg_1(t)]_k=\sum_{j=1}^{r}[{\balpha}_k(t)]_{j}[\nabla f_j(\bw_j(t))]_k$. Here, ${\balpha}_k(t)$ is an element of the probability simplex associated with the $k$-th coordinate of the consensus vector $\bv(t)$, as described earlier. Next, define another vector $\bg_2(t)$ whose $k$-th entry is defined in terms of gradients of the local loss functions evaluated at the consensus vector as $[\bg_2(t)]_k=\sum_{j=1}^{r}[\balpha_k(t)]_{j}[\nabla f_j(\bv(t))]_k$. Further, define a new sequence $\bu(t+1)$ as
\begin{align}
    \bu(t+1)=\bv(t)-\rho(t)\bg_2(t).
\end{align}
It then follows from \eqref{eqn: consensus in one dim}, Assumption~\ref{assumption lipschitz}, and some algebraic manipulations that
\begin{align}
a_3(t) &= \|\bv(t+1)-\bu(t+1)\| = \rho(t) \|\bg_2(t)-\bg_1(t)\| \nonumber\\&\leq \rho(t)L'\max_{j\in \cR}\|\bv(t)-\bw_j(t)\|={\rho(t) L' a_4(t)}.
\end{align}

We next turn our attention to a bound on $a_2(t)$, which is necessarily going to be probabilistic in nature because of its dependence on the training samples, and define a new sequence $\bx(t)$ as
\begin{align}
    \bx(t+1)=\bv(t)-\rho(t)\nabla \E_{\bbP}[f(\bv(t),\bz)].
\end{align}
Trivially, we have
\begin{align}
    a_2(t) &= \|\bu(t+1)-\bx(t+1)\|\nonumber \\&=\rho(t) \|\bg_2(t)-\E_{\bbP}[f(\bv(t),\bz)]\|.
\end{align}
The following lemma, whose proof is provided in Appendix~\ref{appendix: prove of a2}, now bounds $a_2(t)$ by establishing that $\bg_2(t)$ converges in probability to $\E_{\bbP}[f(\bv(t),\bz)]$.
\begin{lemma}\label{lemma gradient converges}
Suppose Assumptions \ref{assumption lipschitz} and \ref{assumption finite value} are satisfied and the training data are i.i.d. Then, fixing any $\delta \in (0,1)$, we have with probability at least $1-\delta$ that
\begin{align}
a_2(t)&\leq \rho(t) \sup_{t} \| \bg_2(t) - \E_{\bbP}[f(\bv(t),\bz)]\| \nonumber \\&= \cO\bigg(\sqrt{\frac{d\|{{\balpha_m}}\|^2\log\frac{2}{\delta}}{N}}\bigg)\rho(t),
\end{align}
where the vector ${\balpha_m}\in\R^{r}$ is a problem-dependent (unknown) vector defined in Appendix \ref{appendix: prove of a2} and satisfies $[{\balpha_m}]_j\geq 0$ and $\sum_{j=1}^{r}[{\balpha_m}]_j = 1$.
\end{lemma}

Notice that while the bounds for $a_2(t)$, $a_3(t)$, and $a_4(t)$ have been obtained for both the convex and nonconvex loss functions in the same manner, the bound for $a_1(t) = \| \bx(t+1)-\bw^*\|$ in the nonconvex setting does require the aid of Assumption \ref{assumption positive definite hessian}, as opposed to Assumption \ref{assumption strongly convex} in the convex setting. This leads to separate proofs of the final statistical optimality results under Assumption~\ref{assumption strongly convex} and Assumption~\ref{assumption positive definite hessian}.

\subsubsection{Statistical optimality for the convex case} Notice that $\bx(t+1)$ is obtained from $\bv(t)$ by taking a regular gradient descent step, with step size $\rho(t)$, with respect to the gradient $\E_{\bbP}[f(\bv(t),\bz)]$ of the statistical risk. Under Assumptions~\ref{assumption lipschitz} and \ref{assumption strongly convex}, therefore, it follows from our understanding of the behavior of gradient descent iterations that \cite[Chapter 2.1.5]{Nesterovbook}
\begin{align}
    a_1(t)=\|\bx(t+1)-\bw^*\|&\leq  (1-L'\rho(t))\|\bv(t)-\bw^*\|\nonumber \\&\leq (1-\lambda\rho(t))\|\bv(t)-\bw^*\|,
\end{align}
where the last inequality holds because of the fact that $\lambda \leq L'$. We then have the following bound:
\begin{align}\label{eqn: shrink of w_j}
    \|\bv(t+1)-\bw^*\| & \leq (1-\lambda\rho(t))\|\bv(t)-\bw^*\| + a_2(t)
    + a_3(t).
\end{align}

Notice that \eqref{eqn: shrink of w_j} only provides the relationship between steps $t$ and $t+1$. In order to bound the distance $\|\bv(t+1)-\bw^*\|$ in terms of the initial distance $\|\bv(0)-\bw^*\|$, we can recursively make use of \eqref{eqn: shrink of w_j} to arrive at the following lemma.
\begin{lemma}\label{adding iteritively}
Suppose Assumptions \ref{assumption lipschitz}, \ref{assumption finite value}, \ref{assumption strongly convex}, and \ref{assumption reduced graph} are satisfied and the training data are i.i.d. Then fixing any $\delta \in (0,1)$, an upper bound on $\|\bv(t+1)-\bw^*\|$ can be derived with probability at least $1-\delta$ as
\begin{align}\label{eqn:distance equation}
    &\|\bv(t+1)-\bw^*\|\leq \frac{t_0}{t+t_0}C_1+\frac{C_2(N)}{\lambda}+\frac{C_3}{t+t_0}\nonumber \\& \ +C_{4}\frac{1}{t+t_0}\bigg(\frac{1}{t_0}+\frac{1}{1+t_0}+\frac{1}{2+t_0}+\cdots+\frac{1}{t+t_0}\bigg),
\end{align}
where $C_1=\|\bv(0)-\bw^*\|$, $C_2(N)=\cO\Big(\sqrt{\frac{d\|{\balpha_m}\|^2\log\frac{2}{\delta}}{N}}\Big)$, $C_3=\frac{\revise{\sqrt{d}}L'rC_w}{\lambda\Big(1-\mu^\frac{1}{\nu}\Big)}$ and  $C_4=\frac{\revise{\sqrt{d}}LL'r\mu^\frac{1}{\nu}}{t_0\lambda^2\Big(1-\mu^\frac{1}{\nu}\Big)}$.
\end{lemma}
Proof of Lemma~\ref{adding iteritively} is provided in Appendix \ref{appendix:proof of adding iteritively}. Lemma~\ref{adding iteritively} establishes that $\|\bv(t+1)-\bw^*\|$ can be upper bounded by a sum of terms that can be made arbitrarily small for sufficiently large $t$ and $N$. Since $\max_{j \in \cR}\| \bw_j(t)-\bv(t)\|$ can also be made arbitrarily small when $t$ is sufficiently large, we can therefore bound $\|\bw_j(t+1)-\bw^*\|, j \in \cR,$ using the bounds on $\|\bv(t+1)-\bw^*\|$ and $\max_{j \in \cR}\| \bw_j(t+1)-\bv(t+1)\|$ to arrive at the following lemma.

\begin{lemma}\label{lemma:condition for convergence}
Suppose Assumptions \ref{assumption lipschitz}, \ref{assumption finite value}, \ref{assumption strongly convex}, and \ref{assumption reduced graph} are satisfied and the training data are i.i.d. Then fixing any $\delta \in (0,1)$ and any $\epsilon > \frac{C_2(N)}{\lambda}>0$, we can always find a $t_1$ such that for all $t\geq t_1$ and $j\in \cR$, with probability at least $1-\delta$, $\|\bw_j(t+1)-\bw^*\| \leq \epsilon$.
\end{lemma}

Proof of Lemma \ref{lemma:condition for convergence} is provided in Appendix \ref{appendix: proof of lemma: condition for convergence}. Notice that given a sufficiently large $N$, $C_2(N)$ is arbitrarily small, which means that the iterates of non-Byzantine nodes can be made arbitrarily close to $\bw^*$. We are now ready to state the main result concerning the statistical convergence of BRIDGE-T at the nonfaulty nodes to the global statistical risk minimizer $\bw^*$.

\begin{theorem}\label{theorem convergence}
Suppose Assumptions \ref{assumption lipschitz}, \ref{assumption finite value}, \ref{assumption strongly convex}, and \ref{assumption reduced graph} are satisfied and the training data are i.i.d. Then the iterates of BRIDGE-T  converge sublinearly in $t$ to the minimum of the global statistical risk at each nonfaulty node. In particular, given any $\epsilon>\frac{\epsilon''}{\lambda}>0$, $\forall j\in \cR$, with probability at least $1-\delta$ and for large enough $t$,
\begin{align}\label{eqn: order of w_j 1}
\|\bw_j(t+1)-\bw^*\| \leq  \epsilon,
\end{align}
where $\delta= 2\exp\Big(-\frac{4rN{\epsilon''}^2}{16L^2rd\|{\balpha_m}\|^2+{\epsilon''}^2}+ r\log\Big(\frac{12L\sqrt{rd}}{\epsilon''}\Big) + d\log\Big( \frac{12L'\beta\sqrt{d}}{\epsilon''}\Big)  \Big)$ and $\epsilon''=C_2(N)=\cO\Big(\sqrt{\frac{d\|{\balpha_m}\|^2\log\frac{2}{\delta}}{N}}\Big)$.
\end{theorem}
Proof of Theorem \ref{theorem convergence} is provided in Appendix \ref{appendix: Proof of theorem convergence}. Note that when $N\to\infty$ and when $\rho(t)$ is chosen as a $\cO(1/t)$ sequence, \eqref{eqn: order of w_j 1} leads to a sublinear convergence rate, as shown in Appendix \ref{appendix: Proof of theorem convergence}. Thus, both the algorithmic and statistical convergence rates derived for BRIDGE-T match the existing Byzantine-resilient rates in the decentralized setting~\cite{yang2019byrdie}. \revise{In terms of the statistical guarantees, recall that when there is no failure in the network, the non-resilient gradient descent algorithms such as DGD usually have the statistical learning rate as $\cO\big(\sqrt{1/MN}\big)$, while if each node runs centralized algorithm with the given $N$ samples, the learning rate is given as $\cO\big(\sqrt{1/N}\big)$. BRIDGE-T achieves the statistical learning rate of $\cO\big(\sqrt{\|{\balpha_m}\|^2/N}\big)$, which lies between the rate of centralized learning and DGD. In particular, compared to local learning, BRIDGE-T reduces the sample complexity by a factor of $\|{\balpha_m}\|^2$ for each node by cooperating over a network, but it cannot approach the fault-free rate. This shows the trade-off between sample complexity and robustness.}

\subsubsection{Statistical optimality for the nonconvex case}\label{nonconvex analysis}
A general challenge for optimization methods in the nonconvex setting is the presence of multiple stationary points within the landscape of the loss function. Distributed frameworks in general and potential Byzantine failures within the network in particular make this an even  more challenging problem. We overcome this challenge by making use of Assumption~\ref{assumption positive definite hessian} (local strong convexity) and aiming for local convergence guarantees.

In terms of specifics, recall the positive constant $\beta$ from Assumption~\ref{assumption positive definite hessian} that describes the region of local strong convexity around a stationary point $\bw^*$, let $\beta_1 \leq \beta$ be another positive constant that will be defined shortly, and pick any $\beta_0 \in (0, \beta - \beta_1]$. Then our local convergence guarantees are based on the assumption that $\forall j \in \cR, \bw_j(0)$'s are initialized such that $\bv(0) \in \B(\bw^*,\beta_0)$, with one such choice of initialization being that the $\bw_j(0)$'s at the nonfaulty nodes are initialized within $\B(\bw^*,\beta_0)$. In particular, assuming $\beta \geq \Gamma$ for $\Gamma$ defined in the beginning of Section~\ref{sec: theoretical analysis}, we obtain the following lemma that establishes the boundedness of the iterates $\bw_j(t)$ for any $j\in \cR$ and $t\in \R$, characterizes the relationship between $\beta, \beta_0,$ and $\beta_1$, and helps us understand how large $\beta$ need be as a function of the different parameters.
%
%
%
\begin{lemma}\label{lemma:condition for not going out}
Suppose Assumptions~\ref{assumption lipschitz}, \ref{assumption finite value}, \ref{assumption positive definite hessian}, and \ref{assumption reduced graph} are satisfied and the training data are i.i.d. Then with the initialization described above for $\beta \geq \max\{\Gamma, \beta_1\}$ and $\beta_1$ defined as $\beta_1 := \frac{C_2(N)}{\lambda}+\frac{C_3}{t_0}+\frac{C_4}{t_0^2}+C_5$, the $\bw_j(t)$'s will never escape from $\B(\bw^*, \beta)$ for all $j \in \cR, t \in \R$. Here, the constants $C_2, C_3$, and $C_4$ are as defined in Lemma~\ref{adding iteritively}, while the constant $C_5 := \revise{\sqrt{d}}rC_w\mu^\frac{1}{\nu}+\revise{\sqrt{d}}rL\frac{1}{1-\mu^\frac{1}{\nu}}\left[\frac{1}{\lambda t_0}\mu^\frac{1}{\nu}+\frac{1}{\lambda(t_0+1)}\right]$.
\end{lemma}
Lemma \ref{lemma:condition for not going out} is proved in Appendix \ref{appendix: proof of lemma: condition for not going out}. In terms of the implications of this lemma, it first and foremost helps justify the assumption of bounded iterates of the nonfaulty nodes stated in the beginning of Section~\ref{sec: theoretical analysis}. More importantly, however, notice that the constraint $\beta \geq \Gamma$ and the assumption that $\forall j \in \cR,t \in \R^d$, $\|\bw_j(t)-\bw^*\|\leq \Gamma$ have the potential to make Assumption~\ref{assumption positive definite hessian} meaningless for large-enough $\Gamma$. But Lemma \ref{lemma:condition for not going out} effectively helps us characterize the extent of $\Gamma$, i.e., choosing $\Gamma$ to be $C_1+\frac{C_2(N)}{\lambda}+\frac{C_3}{t_0}+\frac{C_4}{t_0^2}+C_5$ is sufficient to guarantee that all iterates of the nonfaulty nodes remain within $\B(\bw^*, \Gamma) \subseteq \B(\bw^*, \beta)$.
%

We conclude with our main result for the nonconvex case, which mirrors that for the convex loss functions.
\begin{theorem}\label{theorem convergence nonconvex}
Suppose Assumptions \ref{assumption lipschitz}, \ref{assumption finite value}, \ref{assumption positive definite hessian} and  \ref{assumption reduced graph} are satisfied and the training data are i.i.d. Then with the earlier described initialization within $\B(\bw^*,\beta_0)$, the iterates of BRIDGE-T converge sublinearly in $t$ to the stationary point $\bw^*$ of the statistical risk at each nonfaulty node. In particular, given any $\epsilon>\frac{\epsilon''}{\lambda}>0$, $\forall j\in \cR$, with probability at least $1-\delta$ and for large enough $t$,
\begin{align}\label{eqn: order of w_j 2}
\|\bw_j(t+1)-\bw^*\| \leq  \epsilon,
\end{align}
where $\delta= 2\exp\Big(-\frac{4rN{\epsilon''}^2}{16L^2rd\|{\balpha_m}\|^2+{\epsilon''}^2}+ r\log\Big(\frac{12L\sqrt{rd}}{\epsilon''}\Big) + d\log\Big( \frac{12L'\beta\sqrt{d}}{\epsilon''}\Big)  \Big)$ and $\epsilon''=C_2(N)=\cO\Big(\sqrt{\frac{d\|{\balpha_m}\|^2\log\frac{2}{\delta}}{N}}\Big)$.

\end{theorem}
Proof of Theorem \ref{theorem convergence nonconvex} is provided in Appendix \ref{appendix: proof of theorem convergence nonconvex}. It can be seen from the appendix that the proof in the convex setting maps to the nonconvex one without much additional work. The reason for this is the new proof technique being used in the convex case, instead of the one utilized in \cite{yang2019bridge}, which guarantees BRIDGE-T converges to a local stationary point for the nonconvex functions described in Assumption \ref{assumption positive definite hessian}.

\revise{
\begin{remark}
The convergence rates derived for BRIDGE-T are a function of the dimension $d$. Such dimension dependence is typical of many results in (centralized) statistical learning theory~\cite{Song2018,Damek2021}. While there is an ongoing effort to obtain dimension-independent rates in statistical learning~\cite{Foster2018,Damek2021}, we leave an investigation of this within the context of Byzantine-resilient decentralized learning for future work.
\end{remark}
}

\section{Numerical Results}\label{section numerical analysis}
The numerical experiments are separated into three parts. In the first part, we run experiments on the MNIST \revise{and CIFAR-10 datasets} using a linear classifier with squared hinge loss, which is a case that fully satisfies all our assumptions for the theoretical guarantees in the convex setting. In the second part, we run experiments on the MNIST \revise{and CIFAR-10 datasets} using a convolutional neural network. By showing that BRIDGE works in this general nonconvex loss function setting, we establish that BRIDGE indeed works for loss functions that satisfy Assumption~\ref{assumption positive definite hessian}. In the third part, we run experiments on the MNIST dataset with non-i.i.d.\ distributions of data across the agents. The purpose of all the experiments is to provide numerical validation of our theoretical results and address the usefulness of our Byzantine-resilient technique on a broad scope (convex, nonconvex, non-i.i.d.\ data) of machine learning problems.

\begin{figure}[t]
    \centering
    \includegraphics[height=6cm]{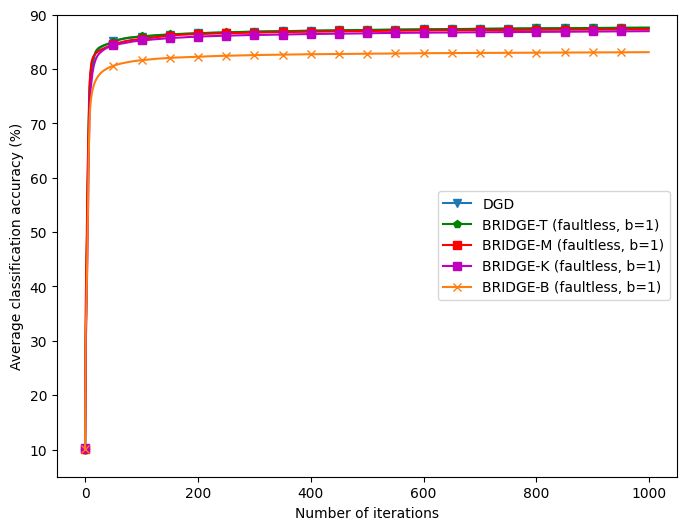}
    \caption{Comparison between DGD and BRIDGE-T, -M, -K, -B in the faultless setting for a convex loss function \revise{and the MNIST dataset, where $b$ is set to be $b=1$ for BRIDGE.}}
    \label{fig:convexfaultless}
    \vspace{-\baselineskip}
\end{figure}

\subsection{Linear classifier on MNIST \revise{and CIFAR-10}}
The first set of experiments is performed to demonstrate two facts: BRIDGE can maintain good performance under Byzantine attacks while classic decentralized learning methods fail; and compared to an existing Byzantine-resilient method, ByRDiE~\cite{yang2019byrdie}, BRIDGE is more efficient in terms of communications cost. We choose one of the most well-understood machine learning tools, the linear classifier with squared hinge loss, to learn the model for this purpose. Note that by showing BRIDGE works in this strictly convex and Lispchitz loss function setting means BRIDGE also works for strongly convex loss functions with bounded Lipschitz gradients.

\looseness=-1 The MNIST dataset is a set of 60,000 training images and 10,000 test images of handwritten digits from `0' to `9'. Each image is converted to a 784-dimensional vector and we distribute 60,000 images equally among 50 nodes. \revise{The CIFAR-10 dataset is a set of 50,000 training images and 10,000 test images of 10 different classes. Each image is converted to a 3072-dimensional vector and we distribute 50,000 images equally among 50 nodes. Then, unless stated otherwise, we connect each pair of nodes with probability $p=0.5$}. Some of the nodes are randomly picked to be Byzantine nodes, which broadcast random vectors to all their neighbors during each iteration. \revise{The parameter $b$ for BRIDGE is set to be $b=1$ in the faultless setting, while it is set to be equal to $|\cB|$ in the faulty setting. Once a random network is generated and the Byzantine nodes are randomly placed, we check for each variant of BRIDGE whether the minimum neighborhood-size condition listed in Table~\ref{table:12} for its execution is satisfied before running that variant.} The classifiers are trained using the ``one vs all" strategy. \revise{We run five sets of experiments, with the first four on the MNIST dataset and the last one on the CIFAR-10 dataset: ($i$) classic distributed gradient descent (DGD) and BRIDGE-T, -M, -K, -B with no Byzantine nodes; ($ii$) classic DGD and BRIDGE-T, -M, -K, -B with 2 and 4 Byzantine nodes; ($iii$) BRIDGE-T, -M, and -K with 6, 12, 18, and 24 Byzantine nodes and varying probabilities of connection ($p=0.5, 0.75$, and $1$); ($iv$) ByRDiE and BRIDGE-T with 2 Byzantine nodes; and ($v$) BRIDGE-T, -M, and -K with 0, 2, 4, and 6 Byzantine nodes.} The performance is evaluated by two metrics: classification accuracy on the 10,000 test images and whether consensus is achieved. When comparing ByRDiE and BRIDGE, we compare the accuracy with respect to the number of communication iterations, which is defined as the number of scalar-valued information exchanged among the neighboring nodes.

\begin{figure}[t]
    \centering
    \includegraphics[height=6cm]{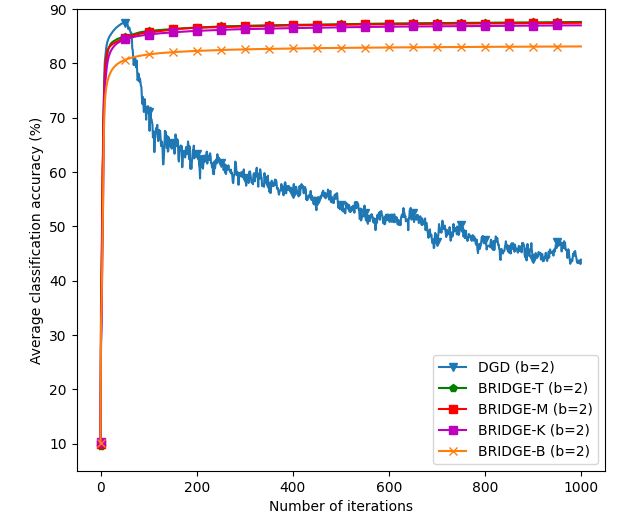}\\
    \vspace{0.5\baselineskip}
    \includegraphics[height=6cm]{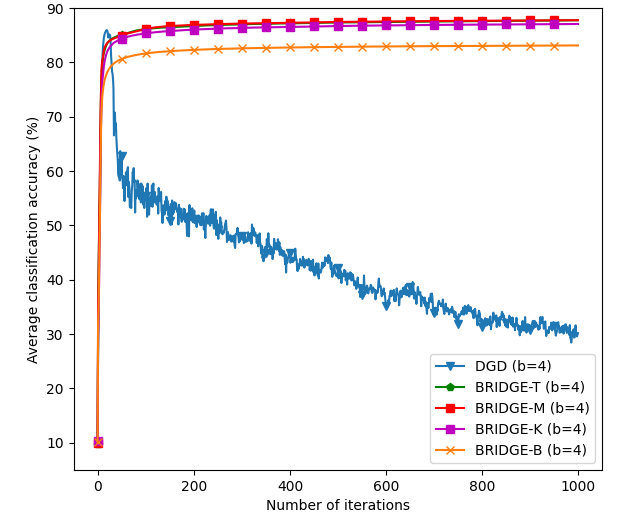}
    \caption{Comparison between DGD and BRIDGE-T, M, K, B with two and four Byzantine nodes for a convex loss function \revise{with the MNIST dataset.}}
    \label{fig:convexfaulty}
    \vspace{-\baselineskip}
\end{figure}

\begin{figure}[t]
    \centering
    \includegraphics[height=6.25cm]{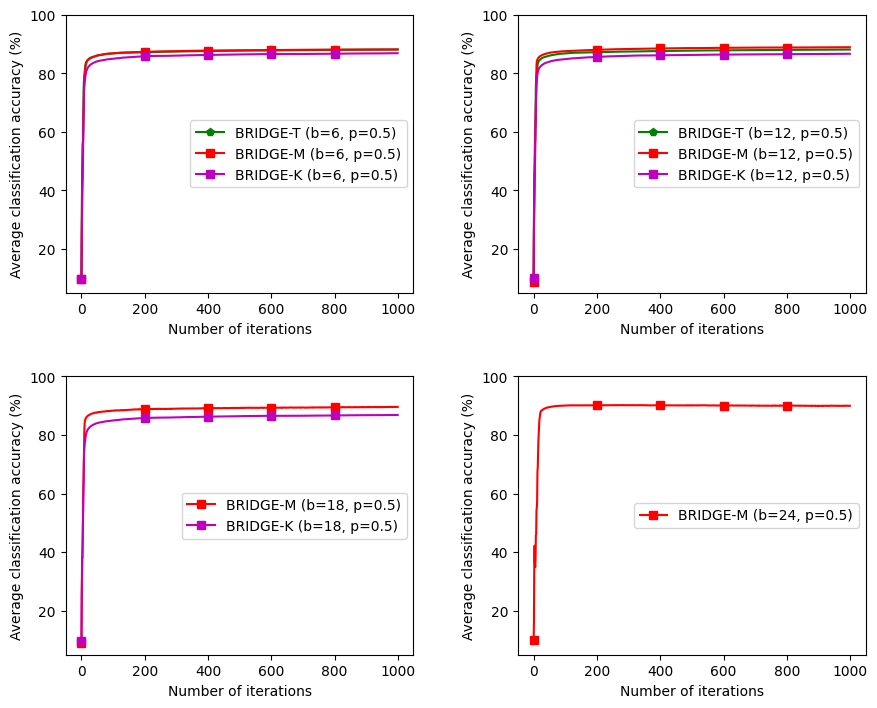}\\
    \vspace{0.5\baselineskip}
    \includegraphics[height=6.25cm]{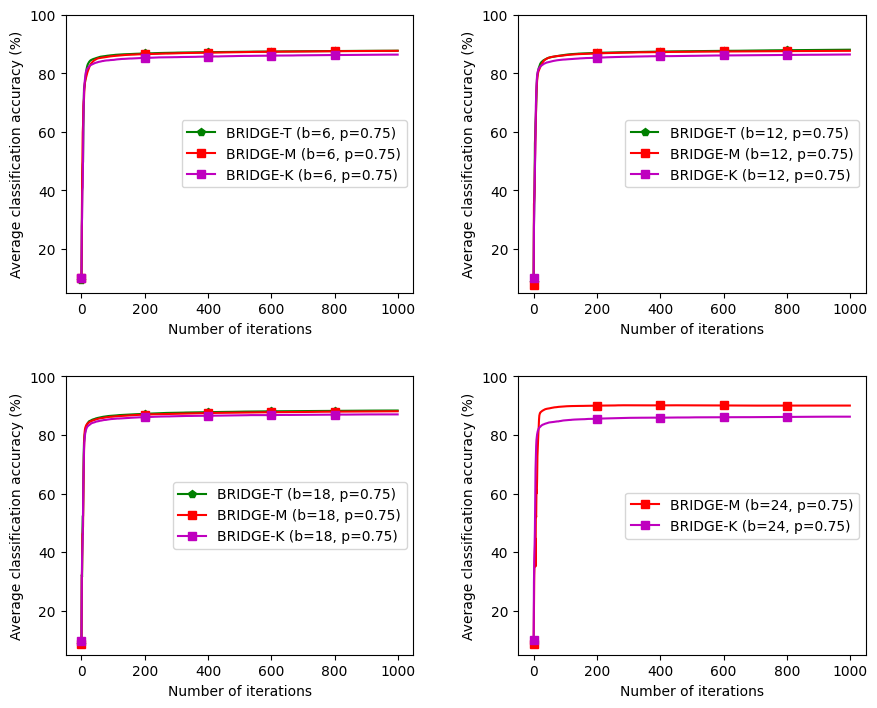}\\
    \vspace{0.5\baselineskip}
    \includegraphics[height=6.25cm]{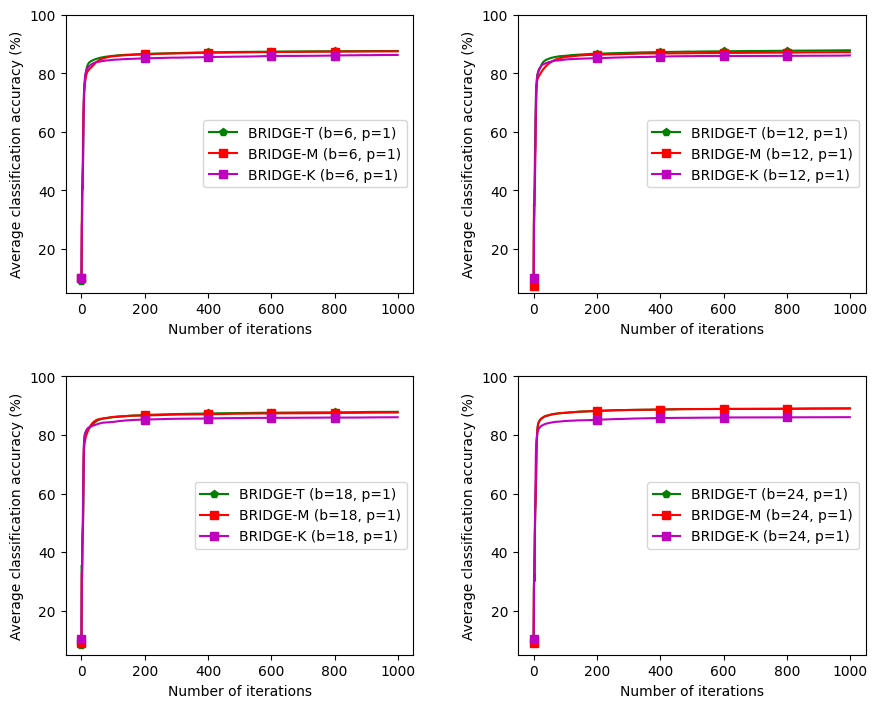}\\
    \caption{\revise{Comparison between BRIDGE-T, -M, and -K for different numbers of Byzantine nodes and varying levels of network connectivity (convex loss and MNIST dataset).}}
    \label{fig:convexmorefaulty}
    \vspace{-\baselineskip}
\end{figure}

\begin{figure}[t]
    \centering
    \includegraphics[height=6cm]{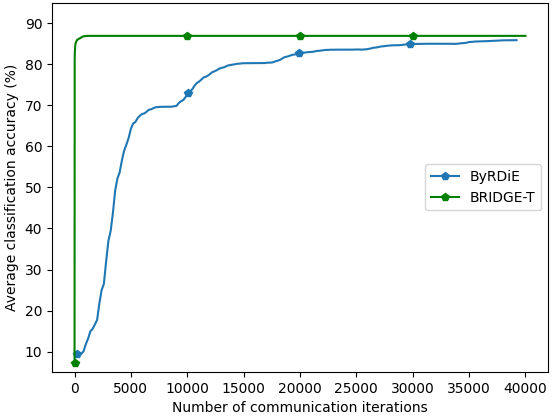}
    \caption{Comparing BRIDGE-T with ByRDiE in the presence of two Byzantine nodes \revise{(convex loss and MNIST dataset).}}
    \label{fig:bvb}
\end{figure}

\begin{figure}[t]
    \centering
    \includegraphics[height=7cm]{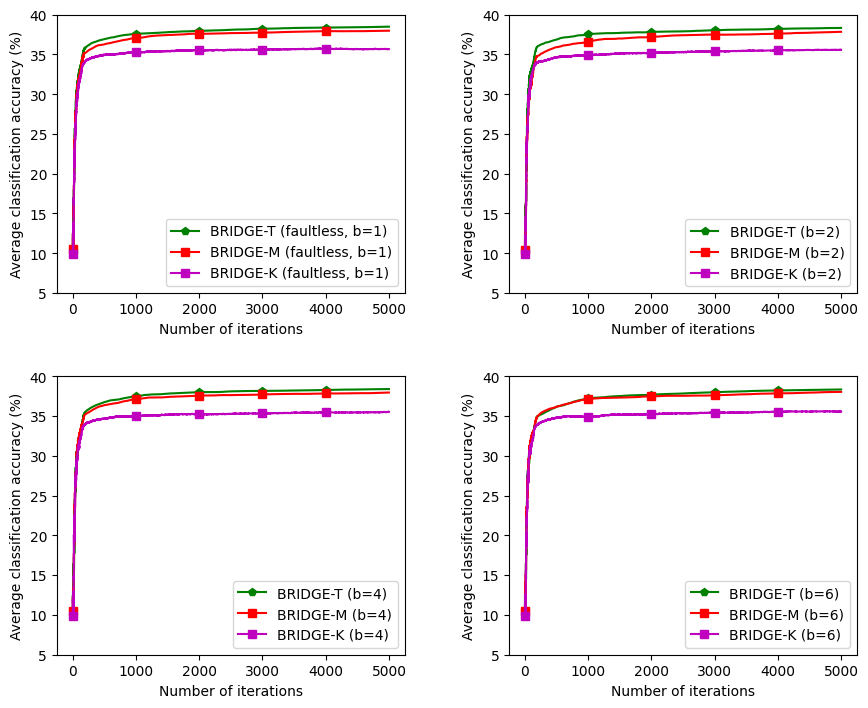}
    \caption{\revise{Performance of BRIDGE-T, -M, and -K with zero, two, four, and six Byzantine nodes for a convex loss function with the CIFAR-10 dataset.}}
    \label{fig:CIFARconvex}
    \vspace{-\baselineskip}
\end{figure}

As we can see from Figure~\ref{fig:convexfaultless}, \revise{despite using $b=1$,} the performances of all BRIDGE methods except BRIDGE-B are as good as DGD in the faultless setting with $\sim 88\%$ average accuracy, while BRIDGE-B performs slightly worse at $83\%$ average accuracy (\emph{final accuracy:} $\text{DGD} = 87.8\%$, $\text{BRIDGE-T} = 87.6\%$, $\text{-M} = 87.3\%$; $\text{-K} = 86.9\%$, and $\text{-B} = 83.1\%$). \revise{Note that these accuracy figures match the state-of-the-art results for the MNIST dataset using a linear classifier~\cite{Lecun1998}. We also attribute the superiority of BRIDGE-T over -M, -K, and -B to its ability to retain information from a wider set of neighbors after the screening in each iteration.} Next, we conclude from Figure~\ref{fig:convexfaulty} that DGD fails in the faulty setting when $|\cB|=2$ and produces an even worse accuracy in the faulty setting when $|\cB|=4$. However, BRIDGE-T, -M, -K and -B with $b = |\cB|$ are able to learn relatively good models in these faulty settings.

\revise{The next set of results in Figure~\ref{fig:convexmorefaulty} highlights the robustness of the BRIDGE framework to a larger number of Byzantine nodes for varying levels of network connectivity that range from $p=0.5$ to $p=1$. We exclude BRIDGE-B in this figure since it fails to run for a majority of the randomly generated networks because of the stringent minimum neighborhood size condition (cf.~Table~\ref{table:12}). The results in this figure reaffirm our findings from Figure~\ref{fig:convexfaulty} that the BRIDGE framework is extremely resilient to Byzantine attacks in the network. In particular, we see that BRIDGE-T (when it satisfies the minimum neighborhood size condition) and -M perform very similarly in the face of a large number of Byzantine nodes in the network, while BRIDGE-K is a close third in performance. Another observation from this figure is the robustness of BRIDGE-M for loosely connected Erd\H{o}s--R\'{e}nyi networks and a large number of Byzantine attacks; indeed, BRIDGE-M is the only variant that can be run in the case of $b=24$ and $p=0.5$ since it always satisfies the minimum neighborhood size condition even when close to $50\%$ of the nodes in the network are Byzantine. In contrast, BRIDGE-T could not be run when $b \geq 18$ (resp., $b=24$) and $p=0.5$ (resp., $p=0.75$), while BRIDGE-K could not be run when $b=24$ and $p=0.5$.}

We next compare BRIDGE-T and ByRDiE in Figure~\ref{fig:bvb}. Both BRIDGE-T and ByRDiE are resilient to two Byzantine nodes but since ByRDiE is based on a coordinate-wise screening method, the time it takes to reach the final optimal solution is thousands-fold more than BRIDGE-T. \revise{Indeed, since nodes within the BRIDGE framework compute the local gradients and communicate with their neighbors only once per iteration, this leads to massive savings in computation and communications costs in comparison with ByRDiE.} This \revise{difference} in terms of computation and communications costs is even more pronounced in higher-dimensional tasks; thus we will not compare with ByRDiE for the rest of our experiments.

\revise{Last but not the least, Figure~\ref{fig:CIFARconvex} highlights the robustness of the BRIDGE framework on the higher-dimensional CIFAR-10 dataset for the case of a linear classifier. It can be seen from this figure that the earlier conclusions concerning the different variants of BRIDGE still hold, with BRIDGE-T approaching the state-of-the-art accuracy of $\sim 39\%$~\cite{le2016} for a linear classifier. We conclude by noting that BRIDGE-B is excluded here and in the following for the experiments on CIFAR-10 dataset because of its relatively high computational complexity on larger-dimensional data.}

\begin{figure}[t]
    \centering
    \includegraphics[height=7cm]{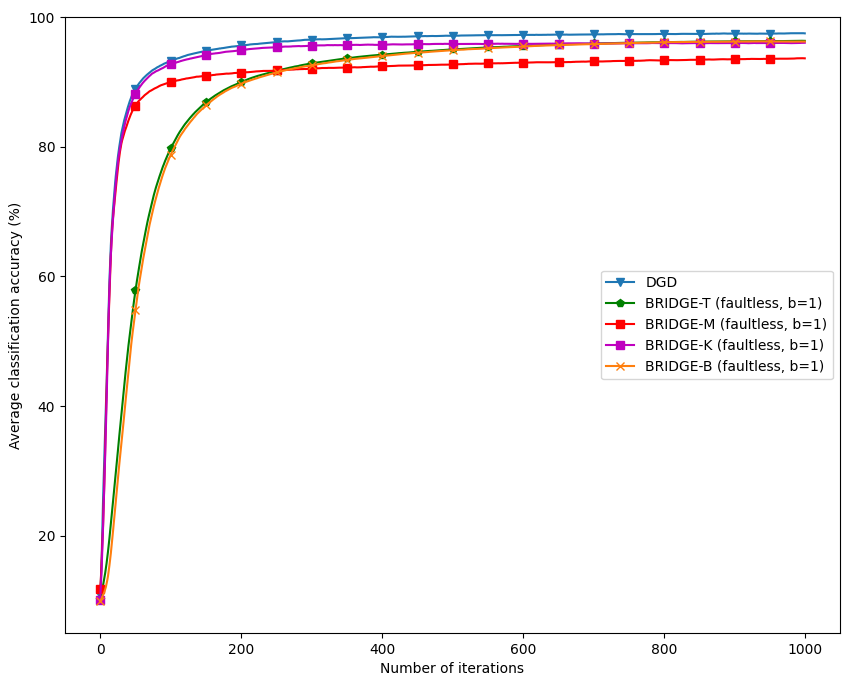}
    \caption{Comparison between DGD and BRIDGE-T, -M, -K, and -B in the faultless setting for a nonconvex loss function \revise{with the MNIST dataset.}}
    \label{fig:nonconvexfaultless}
    \vspace{-\baselineskip}
\end{figure}

\subsection{Convolutional neural network on MNIST \revise{and CIFAR-10}}
In the theoretical analysis, we gave local convergence guarantees of BRIDGE-T in the nonconvex setting. In this set of experiments, we numerically show that BRIDGE indeed performs well in the nonconvex case. We train a convolutional neural network (CNN) on MNIST \revise{and CIFAR-10 datasets} for this purpose, with the model including two convolutional layers followed by two fully connected layers. Each convolutional layer is followed by a max pooling and ReLU activation while the output layer uses softmax activation. We construct a network with 50 nodes and each pair of nodes has probability of 0.5 to be directly connected. Each node has access to 1200 samples randomly picked from the training set. We randomly choose two or four of the nodes to be Byzantine nodes, which broadcast random vectors to all their neighbors during each iteration for all screening methods. We again use the averaged classification accuracy on the MNIST \revise{and CIFAR-10 test sets} over all nonfaulty nodes as the metric for performance. In this experiment, we cannot compare with ByRDiE due to the fact that the learning model is of high dimension, which makes ByRDiE unfeasible in this setting.

\begin{figure}[t]
    \centering
    \includegraphics[height=6cm]{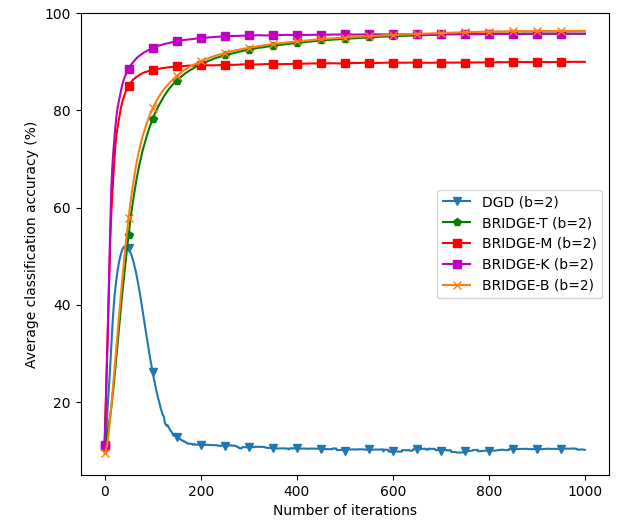}\\
    \vspace{0.5\baselineskip}
    \includegraphics[height=6cm]{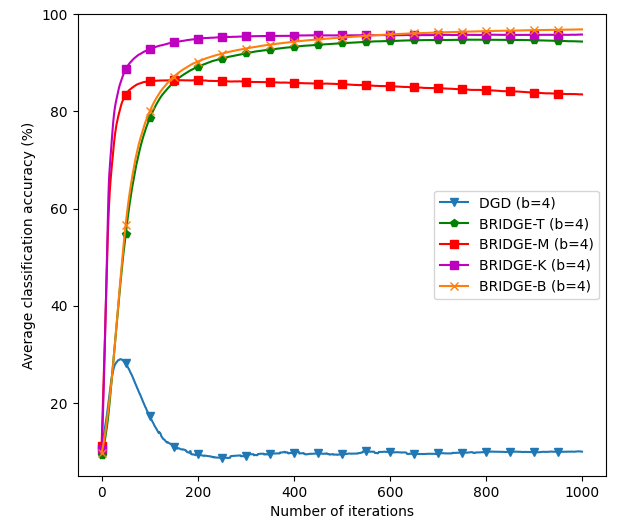}\\
    \caption{Comparison between DGD and BRIDGE-T, -M, -K, and -B with two and four Byzantine nodes for nonconvex loss \revise{with the MNIST dataset.}}
    \label{fig:nonconvexfaulty}
\end{figure}

\begin{figure}[t]
    \centering
    \includegraphics[height=7cm]{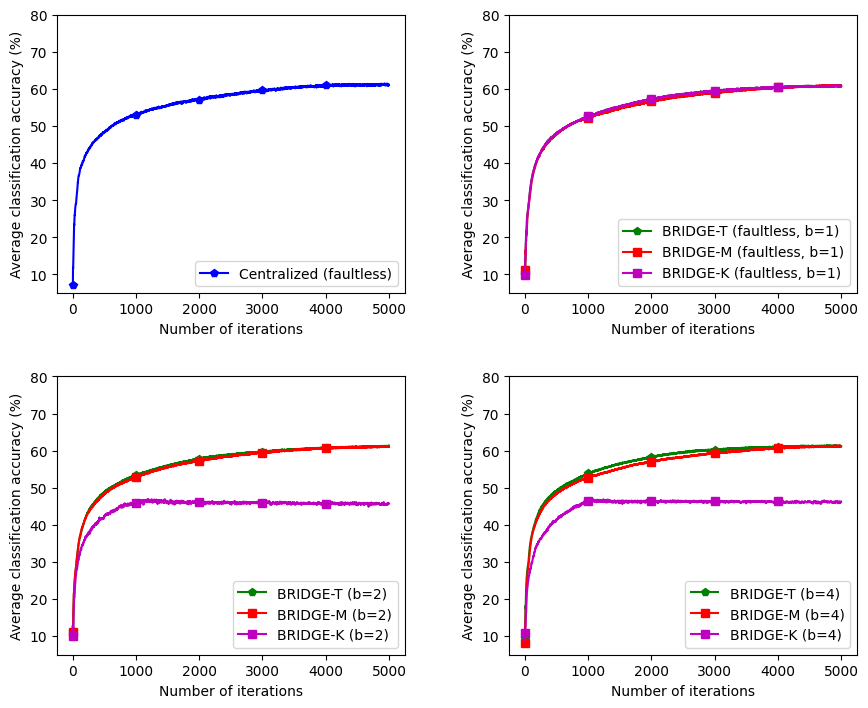}\\
    \vspace{0.5\baselineskip}
    \caption{\revise{Performance comparison of BRIDGE-T, -M, and -K with zero, two, and four Byzantine nodes for nonconvex loss with the CIFAR-10 dataset.}}
    \label{fig:nonconvexfaultyCIFAR}
    \vspace{-\baselineskip}
\end{figure}

As we can see from Figure \ref{fig:nonconvexfaultless}, the performances of all BRIDGE methods are as good as DGD in the faultless setting with 92\% to 95\% percent average accuracy. In Figure \ref{fig:nonconvexfaulty}, we see that DGD fails in the faulty setting when $b=2$ and $b=4$. But BRIDGE-T, -M, -K, and -B are able to learn a relatively good model in these cases. \revise{The final set of results for the CIFAR-10 dataset obtained using BRIDGE-T, -M, and -K for the case of zero, two, and four Byzantine nodes is presented in Figure \ref{fig:nonconvexfaultyCIFAR}. The top-left quadrant in this figure is reserved for the accuracy of the centralized solution for the chosen CNN architecture.}\footnote{\revise{Note that a better CIFAR-10 accuracy can be obtained through a fine tuning of the CNN architecture and the step size sequence.}} \revise{It can be seen that both BRIDGE-T and -M remain resilient to Byzantine attacks and achieve accuracy similar to the centralized solution. However, BRIDGE-K gets stuck in a suboptimal critical point of the loss landscape for the case of two and four Byzantine nodes.}

\subsection{Non-i.i.d.\ data distribution on MNIST}
In the theoretical analysis section, we gave convergence gurantees for BRIDGE-T in both convex and nonconvex settings. However, the main results are based on identical and independent distribution (i.i.d.) of the dataset. In this section, we compare our method to the one proposed in \cite{Peng2020ByzantineRobustDS}, which we term ``Byzantine-robust decentralized stochastic optimization'' (BRDSO) in the following discussion based on the terminology used in \cite{Peng2020ByzantineRobustDS}. We compare BRIDGE-T with BRDSO \cite{Peng2020ByzantineRobustDS} in the following non-i.i.d.\ settings.

\begin{figure}[t]
    \centering
    \includegraphics[height=6cm]{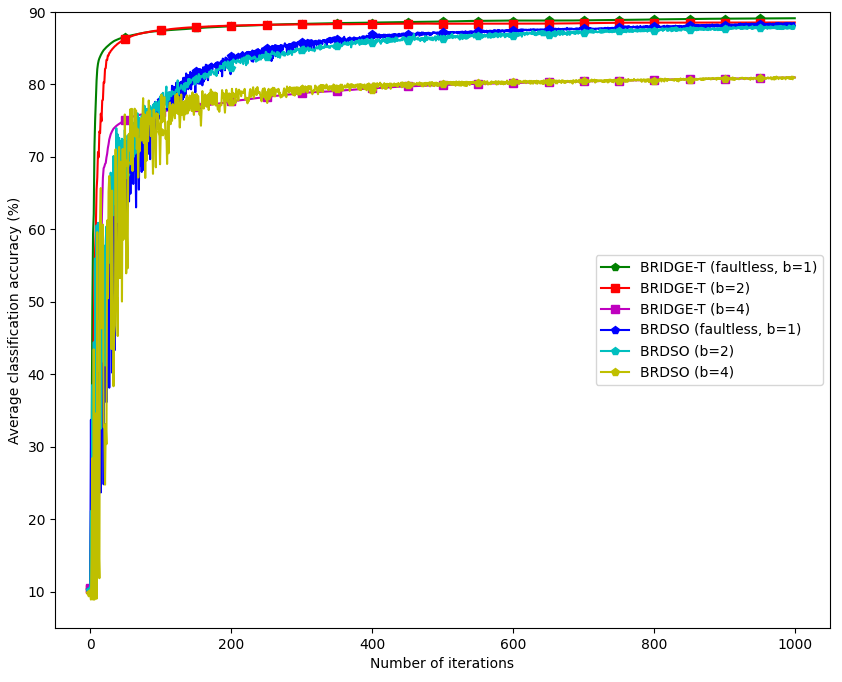}
    \caption{Comparing BRIDGE-T with BRDSO~\cite{Peng2020ByzantineRobustDS} in the extreme non-i.i.d.\ setting (convex loss and MNIST dataset).}
    \label{fig:convexexnoniid}
\end{figure}

\textbf{Extreme non-i.i.d.\ setting:} We group the dataset corresponding to labels and distribute all the samples labelled ``0'' to 5 agents, distribute all the samples labelled ``1'' to another 5 agents, and so on. We can see from Figure \ref{fig:convexexnoniid} that when the number of Byzantine nodes is equal to 0 or 2, the accuracies of both the algorithms are as good as in the i.i.d.\ case, while in the case when the number of Byzantine nodes is equal to 4, there is about 9 percent of accuracy drop due to the non-i.i.d.\ distribution of data for both algorithms. In this extreme non-i.i.d.\ setting when four of the agents are chosen to be Byzantine, the worst-case scenario happens when all the Byzantine nodes are assigned all samples with the same label. This means 80 percent of the samples of one label is not being used towards the training process, which causes both algorithms to underperform compared to the i.i.d.\ setting.

\begin{figure}[t]
    \centering
    \includegraphics[height=6cm]{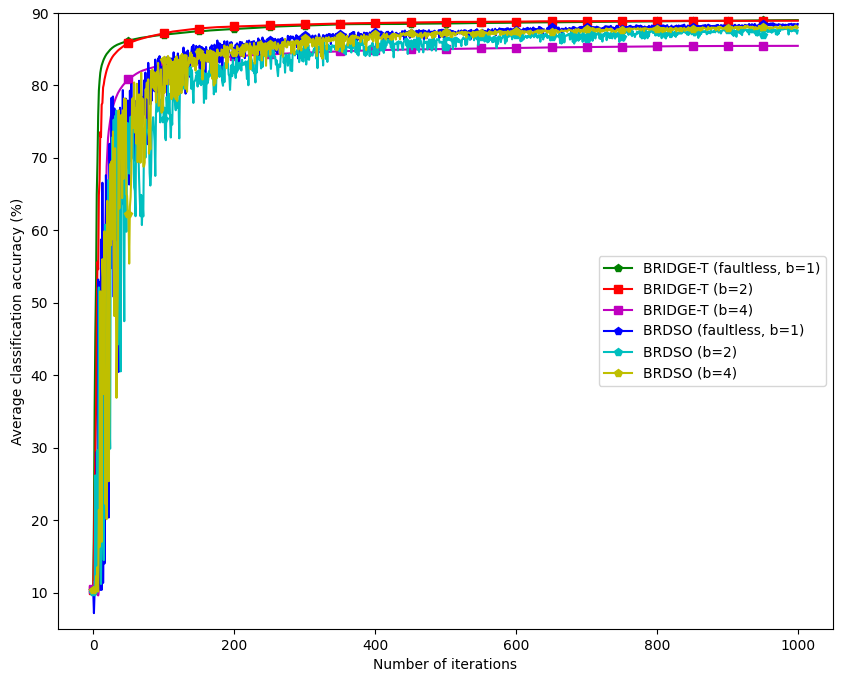}
    \caption{Comparing BRIDGE-T with BRDSO\cite{Peng2020ByzantineRobustDS} in the moderate non-i.i.d.\ setting (convex loss and MNIST dataset)}
    \label{fig:convexmnoniid}
    \vspace{-\baselineskip}
\end{figure}

\textbf{Moderate non-i.i.d.\ setting:} We group the dataset corresponding to labels and distribute the samples associated with each label evenly to 10 agents. Every agent receives two sets of differently labelled data evenly. As we can see from Figure \ref{fig:convexmnoniid}, both algorithms perform as well as in the i.i.d.\ setting in the presence of two or four Byzantine nodes. We conclude that, with distribution closer to i.i.d., the impact of Byzantine nodes in the non-i.i.d.\ setting will be less.

\section{Conclusion}\label{section conclusion}
This paper introduced a new decentralized machine learning framework called Byzantine resilient decentralized
gradient descent (BRIDGE). This framework was designed to solve machine learning problems when the training
set is distributed over a decentralized network in the presence of Byzantine failures. Both theoretical results and experimental results were used to show that the framework could perform well while tolerating Byzantine attacks. One variant of the framework was shown to converge sublinearly to the global minimum in the convex setting and first-order stationary point in the convex setting. In addition, statistical convergence rates were also provided for this variant.

Future work aims to improve the framework to tolerate more Byzantine agents within the network with faster convergence rates and to deal with more general nonconvex objective functions with either i.i.d.\ or non-i.i.d.\ distribution of the dataset. In addition, an exploration of the number of Byzantine nodes that this framework can tolerate under different kinds of non-i.i.d.\ settings will also be undertaken in future work.

\appendix

\subsection{Proof of Lemma \ref{lemma gradient converges}}\label{appendix: prove of a2}

First, we drop $\bbP$ and $\bz$ in notation of the statistical risk for convenience and observe that for any dimension $k$ we have
\begin{align*}
\E[\bg_2(t)]_k=\E\sum\limits_{j=1}^{r}[\balpha_k(t)]_{j}[\nabla f_j(\bv(t))]_k =\E[\nabla f(\bv(t))]_k.
\end{align*}
Since $k$ is arbitrary, it then follows that
\begin{align}
\E[\bg_2(t)]=\E[\nabla f(\bv(t))].
\end{align}
In the definition of $\bg_2(t)$, notice that $\bv(t)$ depends on $t$ and ${\balpha}_k(t)$ depends on both $t$ and $k$. We therefore need to show that the convergence of $\bg_2(t)$ to $\E[\nabla f(\bv(t))]$ is uniformly over all $\bv(t)$ and ${\balpha}_k(t)$. We fix an arbitrary coordinate $k$ and drop the index $k$ for the rest of this section for simplicity. We next define a vector $\bh(t)$ as $\bh(t):=[[\nabla f_j(\bv(t))]:j\in \cR]$ and note that $\bg_2(t)=\langle{\balpha}(t), \bh(t)\rangle$. Since the training data are i.i.d.,  $\bh(t)$ has identically distributed elements. We therefore have from Hoeffding's inequality~\cite{hoeffding1963probability} that for any $\epsilon_0\in (0,1)$:
\begin{align}\label{eqn: single iteration probability}
&\bbP \big(\left\vert\langle{\balpha}(t), \bh(t)\rangle - \E[\nabla f(\bv(t))] \right\vert\geq \epsilon_0 \big) \nonumber\\&\hspace{4cm}\leq 2\exp \Big( -\frac{2N\epsilon_0^2}{L^2\|{\balpha}(t)\|^2}\Big).
\end{align}
Further, since the $r$-dimensional vector ${\balpha}(t)$ is an arbitrary element of the standard simplex, defined as
\begin{align}
  \Delta := \{\bq \in \R^{r}: \sum_{j=1}^{r} [\bq]_j = 1 \text{ and } \forall j, [\bq]_j \geq 0\},
\end{align}
the probability bound in \eqref{eqn: single iteration probability} also holds for any $\bq \in \Delta$, i.e.,
\begin{align}\label{eqn: single iteration probability 2}
\bbP\bigg(\left\vert \langle \bq, \bh(t)\rangle - \E[\nabla f(\bv(t))]\right\vert \geq \epsilon_0\bigg) \leq 2\exp\bigg(-\frac{2N{\epsilon_0}^2}{L^2 \|\bq\|^2}\bigg).
\end{align}

We now define the set $\cS_{\balpha} := \{{\balpha}_k(t)\}_{t,k=1}^{\infty,d}$. Our next goal is to leverage \eqref{eqn: single iteration probability 2} and derive a probability bound similar to \eqref{eqn: single iteration probability} that \emph{uniformly} holds for \emph{all} $\bq \in \cS_{\balpha}$. To this end, let
\begin{align}
  \mathbb{C}_\xi := \{\bc_1,\dots,\bc_{d_\xi}\} \subset \Delta \quad \text{s.t.} \quad \cS_{\balpha} \subseteq \bigcup_{q=1}^{d_\xi} \B(\bc_q, \xi)
\end{align}
denote a $\xi$-covering of $\cS_{\balpha}$ in terms of the $\ell_2$ norm and define $\bar{\bc} := \argmax_{\bc \in \mathbb{\C}_\xi} \|\bc\|$. Then from~\eqref{eqn: single iteration probability 2} and the union bound
\begin{align}\label{eqn:lemma.proof.union.bound.1}
&\bbP\bigg(\sup_{\bc \in \mathbb{C}_\xi} \left\vert \langle\bc, \bh(t)\rangle - \E[\nabla f(\bv(t))]\right\vert \geq \epsilon_0\bigg)\nonumber\\&\hspace{4cm}\leq 2d_\xi \exp\bigg(-\frac{2N{\epsilon_0}^2}{L^2 \|\bar{\bc}\|^2}\bigg).
\end{align}
In addition, we have
\begin{align}\label{eqn:lemma.proof.sup.bounds.1}
  &\sup_{\bq \in \cS_{\balpha}} \left|\langle \bq, \bh(t)\rangle - \E[\nabla f(\bv(t))]\right| \stackrel{(a)}{\leq} \sup_{\bq \in \cS_{\balpha},\bc \in \C_\xi}\|\bq - \bc\| \|\bh(t)\|\nonumber\\&\hspace{2.5cm}+\sup_{\bc \in \bc_\xi} \left|\langle \bc, \bh(t)\rangle - \E[\nabla f(\bv(t))]\right|,
\end{align}
where ($a$) is due to the triangle and Cauchy--Schwarz inequalities. Trivially, $\sup_{\bq \in \cS_{\balpha},\bc \in \C_\xi}\|\bq - \bc\| \leq \xi$ from the definition of $\C_\xi$, while $\|\bh(t)\| \leq \sqrt{r}L$ from the definition of $\bh(t)$ and Assumption~\ref{assumption lipschitz}. Combining \eqref{eqn:lemma.proof.union.bound.1} and \eqref{eqn:lemma.proof.sup.bounds.1}, we get
\begin{align}\label{eqn:lemma.proof.union.bound.2}
\bbP\bigg(\sup_{\bq \in \cS_{\balpha}} \left\vert \langle \bq, \bh(t)\rangle - \E[\nabla f(\bv(t))]\right\vert \geq \epsilon_0 + \sqrt{r}\xi L\bigg)\nonumber\\\leq 2d_\xi \exp\bigg(-\frac{2N{\epsilon_0}^2}{L^2 \|\bar{\bc}\|^2}\bigg).
\end{align}
We now define ${\balpha_m} := \argmax_{\bq \in \cS_{\balpha}} \|\bq\|$. It can then be shown from the definitions of $\mathbb{C}_\xi$ and $\bar{\bc}$ that
\begin{align}\label{eqn:max.alpha.norm}
  \|\bar{\bc}\|^2 \leq 2(\|\balpha_m\|^2 + \xi^2).
\end{align}
Therefore, fixing $\epsilon_0 \in (0,1)$, and defining $\epsilon' := 2\epsilon_0$ and $\xi := \epsilon'/(2L\sqrt{r})$, we have from \eqref{eqn:lemma.proof.union.bound.2} and \eqref{eqn:max.alpha.norm} that
\begin{align}\label{eqn:lemma.proof.union.bound.3}
&\bbP\bigg(\sup_{\bq \in \cS_{\balpha}} \left\vert \langle \bq, \bh(t)\rangle - \E[\nabla f(\bv(t))] \right\vert \geq \epsilon'\bigg)\nonumber\\&\hspace{2.5cm}\leq 2d_\xi \exp\bigg(-\frac{4rN{\epsilon'}^2}{4L^2r\|{\balpha_m}\|^2 + {\epsilon'}^2}\bigg).
\end{align}

Note that \eqref{eqn:lemma.proof.union.bound.3} is derived for a fixed but arbitrary $k$. Extending this to the entire vector gives us for any $\bv(t)$
\begin{align}\label{eqn:lemma.proof.union.bound.4}
&\bbP\bigg( \left\| \bg_2(t) - \E[\nabla f(\bv(t))] \right\| \geq \sqrt{d}\epsilon'\bigg)\nonumber\\&\hspace{2.5cm}\leq 2d_\xi \exp\bigg(-\frac{4rN{\epsilon'}^2}{4L^2r\|{\balpha_m}\|^2 + {\epsilon'}^2}\bigg).
\end{align}
To obtain the desired uniform bound, we next need to remove the dependence on $\bv(t)$ in \eqref{eqn:lemma.proof.union.bound.4}. Here we drop $t$ from $\bv(t)$ for simplicity of notation and write $\bg_2(t)$ as $\bg_2(\bv)$ to show the dependence of $\bg_2$ on $\bv$. Notice from our discussion in the beginning of Section~\ref{sec: theoretical analysis} and the analysis in Section~\ref{ssec:consensus.analysis} that $\bv(t) \in \V := \{\bv: \|\bv\| \leq \Gamma_0 \}$ for some $\Gamma_0$ and all $t$. We then define $\bE_\zeta := \{\be_1,\dots,\be_{m_\zeta}\} \subset \V$ to be a $\zeta$-covering of $\V$ in terms of the $\ell_2$ norm. It then follows from \eqref{eqn:lemma.proof.union.bound.4} that
\begin{align}\label{eqn:lemma.proof.union.bound.5.0}
&\bbP\bigg(\sup_{\be \in \bE_\zeta} \left\| \bg_2(\be) - \E[\nabla f(\be)]\right\| \geq \sqrt{d}\epsilon'\bigg)\nonumber\\&\hspace{2cm}\leq 2d_\xi m_\zeta \exp\bigg(-\frac{4rN{\epsilon'}^2}{4L^2r\|{\balpha_m}\|^2 + {\epsilon'}^2}\bigg).
\end{align}
Similar to \eqref{eqn:lemma.proof.sup.bounds.1}, we can also write
\begin{align}\label{eqn:lemma.proof.sup.bounds.2}
  &\sup_{\bv \in \V} \left\| \bg_2(\bv) - \E[\nabla f(\bv)\|\right\vert\nonumber \leq \sup_{\be \in \bE_\zeta} \left\|\bg_2(\be) - \E[\nabla f(\be)]\right\|\nonumber\\
   &+\sup_{\be \in \bE_\zeta, \bv \in \V}\bigg[\left\|\bg_2(\bv) - \bg_2(\be) \right\| + \left\|\E[\nabla f(\be)]-\E[\nabla f(\bv)]\right\|\bigg].
\end{align}
Further, Assumption~\ref{assumption lipschitz} and definition of the set $\bE_\zeta$ imply
\begin{align}
  \sup_{\be \in \bE_\zeta, \bv \in \V} \left\|\bg_2(\bv) - \bg_2(\be)\right\| &\leq L'\zeta. \label{eqn:lemma.proof.sup.bounds.3}
\end{align}
We now define $\epsilon'' := 2\epsilon'\sqrt{d}$ and $\zeta := \epsilon''/4L'$. We then obtain the following from \eqref{eqn:lemma.proof.union.bound.4}--\eqref{eqn:lemma.proof.sup.bounds.3}:
\begin{align}\label{eqn:lemma.proof.union.bound.5.1}
&\bbP\bigg(\sup_{ \bv\in \V} \left\| \bg_2(\bv) - \E[\nabla f(\bv)]\right\| \geq \epsilon'' \bigg)
\nonumber\\&\hspace{1.5cm}\leq 2d_\xi m_\zeta \exp\bigg(-\frac{4rN {\epsilon''}^2}{16L^2rd\|{\balpha_m}\|^2 + {\epsilon''}^2}\bigg).
\end{align}
Since $\bv(t)\in \V$ for all $t$, we then have
\begin{align}\label{eqn: final union bound}
&\bbP\bigg(\sup_{t} \left\| \bg_2(\bv(t)) - \E[\nabla f(\bv(t))]\right\| \geq \epsilon'' \bigg)
\nonumber\\&\hspace{1.5cm}\leq 2d_\xi m_\zeta \exp\bigg(-\frac{4rN {\epsilon''}^2}{16L^2rd\|{\balpha_m}\|^2 + {\epsilon''}^2}\bigg).
\end{align}

The proof now follows from \eqref{eqn: final union bound} and the following facts about the covering numbers of the sets $\cS_{\balpha}$ and $\V$: (1) Since $\cS_{\balpha}$ is a subset of $\Delta$, which can be circumscribed by a sphere in $\R^{r-1}$ of radius $\sqrt{r-1/r} < 1$, we can upper bound $d_\xi$ by $\Big(\frac{12L\sqrt{rd}}{\epsilon''}\Big)^{r}$~\cite{Verger-Gaugry2005covering}; and (2) Since $\V \subset \R^d$ can be circumscribed by a sphere in $\R^d$ of radius $\Gamma_0 \sqrt{d}$, we can upper bound $m_\zeta$ by $\Big(\frac{12 L' \Gamma_0 \sqrt{d}}{\epsilon''}\Big)^{d}$. Then for any $\epsilon''\in (0,1)$, we have
\begin{align}
\sup_{t} \left\| \bg_2(\bv(t)) - \E[\nabla f(\bv(t))]\right\| < \epsilon''
\end{align}
with probability exceeding
\begin{align}
&1-2\exp\bigg(-\frac{4rN{\epsilon''}^2}{16L^2rd\|{\balpha_m}\|^2+{\epsilon''}^2} + r\log\bigg(\frac{12L\sqrt{rd}}{\epsilon''}\bigg)\nonumber\\&\hspace{4cm}+d\log\bigg( \frac{12L'\Gamma_0\sqrt{d}}{\epsilon''}\bigg)\bigg).
\end{align}
Equivalently, we have with probability at least $1-\delta$ that
\begin{align*}
\sup_{t} \left\| \bg_2(\bv(t)) - \E[\nabla f(\bv(t))]\right\| < \cO\bigg(\sqrt{\frac{4L^2d\|{\balpha_m}\|^2\log\frac{2}{\delta}}{N}}\bigg),
\end{align*}
where $\delta=2\exp\Big(-\frac{4rN{\epsilon''}^2}{16L^2rd\|{\balpha_m}\|^2+{\epsilon''}^2} + r\log\Big(\frac{12L\sqrt{rd}}{\epsilon''}\Big)+ d\log\Big( \frac{12L'\Gamma_0\sqrt{d}}{\epsilon''}\Big)  \Big)$. \hfill $\blacksquare$

\subsection{Proof of Lemma \ref{adding iteritively}}\label{appendix:proof of adding iteritively} 
All statements in this proof are probabilistic, with probability at least $1-\delta$ and $\delta$ being given in Appendix \ref{appendix: prove of a2}. In particular, the discussion should be assumed to have been implicitly conditioned on this high probability event. From \eqref{eqn: shrink of w_j}, we iteratively add up from $\bv(0)$ to $\bv(t+1)$ and get
\begin{align}\label{eqn:add iteritively}
    &\|\bv(t+1)-\bw^*\| \leq \bigg(1-\frac{1}{t+t_0}\bigg)\|\bv(t)-\bw^*\|\nonumber\\&\hspace{4cm}+a_2(t)+a_3(t)\nonumber\\
    &\leq \bigg(1-\frac{1}{t+t_0}\bigg)\bigg[\bigg(1-\frac{1}{t-1+t_0}\bigg)\|\bv(t-1)-\bw^*\|\nonumber\\&\hspace{1cm}+a_2(t-1,N)+a_3(t-1)\bigg]+a_2(t)+a_3(t)\nonumber\\&\leq \prod_{\tau=0}^{t} \bigg(1-\frac{1}{\tau+t_0}\bigg)\|\bv(0)-\bw^*\|+a_2(t)+a_3(t)\nonumber\\&\hspace{1cm}+ \sum_ {\tau=0}^{t-1} [a_2(\tau)+a_3(\tau)]\prod_{z=\tau+1}^{t}\bigg(1-\frac{1}{z+t_0}\bigg).
\end{align}
The first term in \eqref{eqn:add iteritively} can be simplified as
\begin{align}\label{first term}
    &\prod_{\tau=0}^{t} \bigg(1-\frac{1}{\tau+t_0}\bigg)\|\bv(0)-\bw^*\|\nonumber\\&=\bigg(1-\frac{1}{0+t_0}\bigg)\bigg(1-\frac{1}{1+t_0}\bigg)\cdots\bigg(1-\frac{1}{t+t_0}\bigg)\|\bv(0)-\bw^*\|
    \nonumber\\&=\bigg(\frac{t_0-1}{t+t_0}\bigg) C_1,
\end{align}
where $C_1=\|\bv(0)-\bw^*\|$.

The remaining terms in \eqref{eqn:add iteritively} can be simplified as
\begin{align}\label{second term}
    &a_2(t)+a_3(t)+ \sum_ {\tau=0}^{t-1} [a_2(\tau)+a_3(\tau)]\prod_{z=\tau+1}^{t}\bigg(1-\frac{1}{z+t_0}\bigg)\nonumber\\&\leq \bigg(1-\frac{1}{t+t_0}\bigg)a_2(t-1)\nonumber\\&\hspace{0.5cm}+\bigg(1-\frac{1}{t+t_0}\bigg)\bigg(1-\frac{1}{t-1+t_0}\bigg)a_2(t-2)\nonumber\\&\hspace{0.5cm}+a_2(t)+\cdots+\bigg(1-\frac{1}{t+t_0}\bigg)\cdots\bigg(1-\frac{1}{1+t_0}\bigg)a_2(0)\nonumber\\&\hspace{0.5cm}+\frac{L'}{\lambda(t+t_0)}a_4(t)+\bigg(1-\frac{1}{t+t_0}\bigg)\frac{L'}{\lambda(t-1+t_0)}a_4(t-1)\nonumber\\&\hspace{0.5cm}+\cdots+\bigg(1-\frac{1}{t+t_0}\bigg)\cdots\bigg(1-\frac{1}{1+t_0}\bigg)\frac{L'}{\lambda(t_0)} a_4(0)\nonumber\\&=C_2(N)\rho(t)+C_2(N)\rho(t-1)\frac{t+t_0-1}{t+t_0}\nonumber\\&\hspace{0.5cm}+\cdots+C_2(N)\rho(0)\frac{t_0}{t+t_0}+\frac{L'}{\lambda(t+t_0)}a_4(t)\nonumber\\&\hspace{0.5cm}+\frac{t+t_0-1}{\lambda(t+t_0)}\frac{1}{t-1+t_0}L'a_4(t-1)\nonumber\\&\hspace{0.5cm}+\cdots+\frac{t_0}{\lambda(t+t_0)}\frac{1}{t_0}L'a_4(0)\nonumber\\&=\frac{t}{\lambda(t+t_0)}C_2(N)+\frac{L'}{\lambda}\frac{1}{t_0+t}[a_4(0)+\cdots+a_4(t)],
\end{align}
where $C_2(N)=\cO\Big(\sqrt{\frac{d\|{\balpha_m}\|^2\log\frac{2}{\delta}}{N}}\Big)$.\\

The term $a_4(0)+\cdots+a_4(t)$ in \eqref{second term} can be further simplified using \eqref{eqn: consensus in one dim} as:
\begin{align}\label{simplification of a_4 sum}
    &a_4(0)+\cdots+a_4(t)\leq \revise{\sqrt{d}}\bigg(rC_w+rL\frac{1}{\lambda t_0}\mu^\frac{1}{\nu}\bigg)\nonumber\\
    &\quad +\revise{\sqrt{d}}\bigg[rC_w\mu^\frac{1}{\nu}+rL\frac{1}{\lambda t_0}\mu^\frac{2}{\nu}+rL\frac{1}{\lambda (1+t_0)}\mu^\frac{1}{\nu}\bigg]+\cdots\nonumber\\
    &\quad+\revise{\sqrt{d}}\bigg[rC_w\mu^\frac{t}{\nu}+rL\frac{1}{\lambda t_0}\mu^\frac{t+1}{\nu}+rL\frac{1}{\lambda (1+t_0)}\mu^\frac{t}{\nu}\nonumber\\
    &\quad+rL\frac{1}{\lambda (2+t_0)}\mu^\frac{t-1}{\nu}+\cdots+rL\frac{1}{\lambda (t+t_0)}\mu^\frac{1}{\nu}\bigg]\nonumber\\
    &\leq\revise{\sqrt{d}}\bigg(rC_w+rC_w\mu^\frac{1}{\nu}+rC_w\mu^\frac{2}{\nu}+\cdots+rC_w\mu^\frac{t}{\nu}\bigg)\nonumber\\
    &\quad+\frac{\revise{\sqrt{d}}rL}{\lambda}\mu^\frac{1}{\nu}\bigg(\frac{1}{t_0}+\frac{1}{t_0+1}+\cdots\frac{1}{t_0+t}\bigg)\nonumber\\
    &\quad+\frac{\revise{\sqrt{d}}rL}{\lambda}\mu^\frac{2}{\nu}\bigg(\frac{1}{t_0}+\frac{1}{t_0+1}+\cdots+\frac{1}{t_0+t}\bigg)+\cdots\nonumber+\\
    &\quad+\frac{\revise{\sqrt{d}}rL}{\lambda}\mu^\frac{t}{\nu}\bigg(\frac{1}{t_0}+\frac{1}{t_0+1}+\cdots+\frac{1}{t_0+t}\bigg)\nonumber\\
    &=\revise{\sqrt{d}}rC_w\frac{(1-\mu^\frac{t}{\nu})}{1-\mu^\frac{1}{\nu}}\nonumber\\
    &\quad+\frac{\revise{\sqrt{d}}rL}{\lambda}\frac{\mu^\frac{1}{\nu}(1-\mu^\frac{t+1}{\nu})}{1-\mu^\frac{1}{\nu}}\bigg(\frac{1}{t_0}+\frac{1}{t_0+1}+\cdots+\frac{1}{t_0+t}\bigg).
\end{align}
Plugging \eqref{simplification of a_4 sum} into \eqref{second term} we finish the simplification of the remaining terms as:
\begin{align}\label{simplification of second term}
    &a_2(t)+a_3(t)+ \sum_ {\tau=0}^{t-1} [a_2(\tau)+a_3(\tau)]\prod_{z=\tau+1}^{t}\bigg(1-\frac{1}{z+t_0}\bigg)\nonumber
    \\&\leq \frac{1}{\lambda}C_2(N)+\frac{L'}{\lambda (t_0+t)}\bigg(\frac{\revise{\sqrt{d}}rC_w}{1-\mu^\frac{1}{\nu}}\bigg)\nonumber\\
    &\quad+\frac{\revise{\sqrt{d}}LL'r\mu^\frac{1}{\nu}}{\lambda^2\bigg(1-\mu^\frac{1}{\nu}\bigg)}\frac{1}{t+t_0}\bigg(\frac{1}{t_0}+\frac{1}{1+t_0}+\cdots+\frac{1}{t+t_0}\bigg).
\end{align}
Finally we express $ \|\bv(t+1)-\bw^*\|$ using \eqref{first term} and \eqref{simplification of second term} as:
\begin{align}\label{v(t)expression}
     &\|\bv(t+1)-\bw^*\|\leq \frac{t_0}{t+t_0}C_1+\frac{C_2(N)}{\lambda}+\frac{C_3}{t+t_0}\nonumber\\
     &\qquad+\frac{C_4}{t+t_0}\bigg(\frac{1}{t_0}+\frac{1}{1+t_0}+\cdots+\frac{1}{t+t_0}\bigg),
\end{align}
where $C_3=\frac{\revise{\sqrt{d}}rC_wL'}{\lambda\Big(1-\mu^\frac{1}{\nu}\Big)}$ and $C_4=\frac{\revise{\sqrt{d}}LL'r\mu^\frac{1}{\nu}}{\lambda^2\Big(1-\mu^\frac{1}{\nu}\Big)}$. \hfill $\blacksquare$

\subsection{Proof of Lemma \ref{lemma:condition for convergence}}\label{appendix: proof of lemma: condition for convergence}
Similar to the proof of Lemma \ref{adding iteritively}, our statements in this appendix are also conditioned on the high probability event described in Appendix \ref{appendix: prove of a2}.
From Theorem \ref{theorem: consensus a_4 distance} and \eqref{v(t)expression} in the proof of Lemma \ref{adding iteritively}, we conclude for all $j\in\cR$ that
\begin{align}\label{middlestep}
     &\|\bw_j(t+1)-\bw^*\|\leq \frac{t_0}{t+t_0}C_1+\frac{C_2(N)}{\lambda}+\frac{C_3}{t+t_0}\nonumber\\
     &\ +\frac{C_4}{t+t_0}\bigg(\frac{1}{t_0}+\frac{1}{1+t_0}+\cdots+\frac{1}{t+t_0}\bigg)+a_4(t+1).
\end{align}
Next, we get an upper bound on $a_4(t+1)$ as following when we choose $t$ as an even number:
\begin{align}\label{a_4(t) upper bound}
    &a_4(t+1)\leq \revise{\sqrt{d}}rC_w\mu^\frac{t+1}{\nu}+\revise{\sqrt{d}}rL\bigg[\rho(0)\mu^\frac{t+1}{\nu}+\rho(1)\mu^\frac{t}{\nu}+\cdots\nonumber\\
    &\qquad\qquad\qquad+\rho(t-1)\mu^\frac{2}{\nu}+\rho(t)\mu^\frac{1}{\nu}+\rho(t+1)\bigg]\nonumber\\&\hspace{1.4cm}\leq \revise{\sqrt{d}}rC_w\mu^\frac{t+1}{\nu}\nonumber\\
    &\qquad\qquad\qquad+\revise{\sqrt{d}}rL\left[\rho(0)\mu^\frac{t+1}{\nu}+\cdots+\rho(0)\mu^\frac{\frac{t}{2}+1}{\nu}\right]\nonumber\\
    &+\revise{\sqrt{d}}rL\left[\rho\bigg(\frac{t}{2}+1\bigg)\mu^\frac{\frac{t}{2}}{\nu}+\cdots+\rho\bigg(\frac{t}{2}+1\bigg)\mu^\frac{1}{\nu}+\rho\bigg(\frac{t}{2}+1\bigg)\right]\nonumber\\
    &\hspace{1.4cm}= \revise{\sqrt{d}}rC_w\mu^\frac{t+1}{\nu}+\revise{\sqrt{d}}rL\rho(0)\mu^\frac{t+2}{2\nu}\frac{1-\mu^\frac{t}{2\nu}}{1-\mu^\frac{1}{\nu}}\nonumber\\
    &\qquad\qquad\qquad+\revise{\sqrt{d}}rL\rho\bigg(\frac{t}{2}+1\bigg)\frac{1-\mu^\frac{t}{2\nu}}{1-\mu^\frac{1}{\nu}}\nonumber\\
    &\hspace{1.4cm}\leq \revise{\sqrt{d}}rC_w\mu^\frac{t+1}{\nu}\nonumber\\
    &\qquad\qquad\qquad+\revise{\sqrt{d}}rL\frac{1}{1-\mu^\frac{1}{\nu}}\left[\rho(0)\mu^\frac{t+2}{2\nu}+\rho\bigg(\frac{t}{2}+1\bigg)\right].
\end{align}
When $t$ is an odd number, we have:
\begin{align}\label{a_4(t) upper bound 2}
    &\revise{\sqrt{d}}rL\bigg[\rho(0)\mu^\frac{t+1}{\nu}+\rho(1)\mu^\frac{t}{\nu}\nonumber\\&\hspace{1.4cm}+\cdots+\rho(t-1)\mu^\frac{2}{\nu}+\rho(t)\mu^\frac{1}{\nu}+\rho(t+1)\bigg] \nonumber\\&\quad \leq \revise{\sqrt{d}}rL\bigg[\rho(0)\mu^\frac{t+1}{\nu}+\cdots+\rho(0)\mu^\frac{\frac{t}{2}+\frac{1}{2}}{\nu}\bigg]\nonumber\\&\hspace{1.4cm}+\revise{\sqrt{d}}rL\bigg[\rho\bigg(\frac{t}{2}+\frac{1}{2}\bigg)\mu^\frac{\frac{t}{2}-\frac{1}{2}}{\nu}+\cdots+\rho\bigg(\frac{t}{2}+\frac{1}{2}\bigg)\mu^\frac{1}{\nu}\nonumber\\&\hspace{1.4cm}+\rho\bigg(\frac{t}{2}+\frac{1}{2}\bigg)\bigg]
\end{align}
and the remaining steps are similar to \eqref{a_4(t) upper bound}, so we omit them.

Plugging either \eqref{a_4(t) upper bound} or \eqref{a_4(t) upper bound 2} into \eqref{middlestep}, we have for all $j\in\cR$
\begin{align}\label{w(t)upper bound}
     &\|\bw_j(t+1)-\bw^*\|\leq \frac{t_0}{t+t_0}C_1+\frac{C_2(N)}{\lambda}+\frac{C_3}{t+t_0}\nonumber\\&+\frac{C_4}{t+t_0}\bigg(\frac{1}{t_0}+\frac{1}{1+t_0}+\cdots+\frac{1}{t+t_0}\bigg)\nonumber\\&+\revise{\sqrt{d}}rC_w\mu^\frac{t+1}{\nu}+\revise{\sqrt{d}}rL\frac{1}{1-\mu^\frac{1}{\nu}}\left[\rho(0)\mu^\frac{t+2}{2\nu}+\rho\bigg(\frac{t}{2}+1\bigg)\right].
\end{align}
From \eqref{w(t)upper bound} we see that all the terms except $\frac{C_2(N)}{\lambda}$ are monotonically decreasing with increasing $t$. Thus, given any $\epsilon> \frac{C_2(N)}{\lambda}> 0$, we can find a $t_1$ such that for all $t\geq t_1$, with probability at least $1-\delta$, $\|\bw_j(t+1)-\bw^*\|\leq \epsilon$. \hfill $\blacksquare$

\subsection{Proof of Theorem \ref{theorem convergence}}\label{appendix: Proof of theorem convergence}
Lemma \ref{lemma:condition for convergence} establishes the convergence of $\bw_j(t+1)$ to $\bw^*$ for all $j \in \cR$ with probability at least $1-\delta$. Next, we derive the rate of convergence. From \eqref{middlestep} since $a_4(t)$ has a convergence rate of $\cO(1/t)$, as mentioned in Theorem \ref{theorem: consensus a_4 distance}, the term that \revise{converges the} slowest to $0$ among all the terms in $\|\bw_j(t+1)-\bw^*\|$ is $\frac{C_4}{t+t_0}\Big(\frac{1}{t_0}+\frac{1}{1+t_0}+\cdots+\frac{1}{t+t_0}\Big)$.
\revise{Therefore, using the harmonic series approximation, we conclude that the convergence rate is $\cO \left(\frac{\log{t}}{t}\right)$.} The above statement shows BRIDGE-T converges to the minimum of the global statistical risk at a sublinear rate, thus completing the proof of Theorem~\ref{theorem convergence}. \hfill $\blacksquare$

\subsection{Proof of Lemma \ref{lemma:condition for not going out} }\label{appendix: proof of lemma: condition for not going out}
Under the stated assumptions of the lemma as well as the initialization for the nonconvex loss function, it is straightforward to see that \eqref{w(t)upper bound} also holds for the nonconvex setting. In particular, the upper bound in \eqref{w(t)upper bound} on $\|\bw_j(t+1)-\bw^*\|$ for all $j\in\cR$ monotonically decreases with $t$. Thus, $\|\bw_j(t+1)-\bw^*\|$ for all $j\in \cR, t\in\R$ can be upper bounded by the bound on $\|\bw_j(1)-\bw^*\|$, which is $C_1+\frac{C_2(N)}{\lambda}+\frac{C_3}{t_0}+\frac{C_4}{t_0^2}+C_5$. The proof now follows from the fact that $C_1 := \|\bv(0)-\bw^*\| \leq \beta_0$ by virtue of the initialization. \hfill $\blacksquare$

\subsection{Proof of Theorem \ref{theorem convergence nonconvex}} \label{appendix: proof of theorem convergence nonconvex}
The local strong convexity of the loss function implies that $f(\bw,\bz)$ can be treated as $\lambda$-strongly convex when restricted to the ball $\B(\bw^*,\beta)$. Therefore, Assumption~\ref{assumption positive definite hessian}, the $\Gamma$-boundedness of the iterates stated in the beginning of Section \ref{sec: theoretical analysis} and Lemma~\ref{lemma:condition for not going out}, and the constraint $\beta\geq\Gamma$ imply that the proof of this theorem is a straightforward replication of the proof of Theorem \ref{theorem convergence} for the convex case. \hfill $\blacksquare$

\balance


\begin{thebibliography}{10}
\providecommand{\url}[1]{#1}
\csname url@samestyle\endcsname
\providecommand{\newblock}{\relax}
\providecommand{\bibinfo}[2]{#2}
\providecommand{\BIBentrySTDinterwordspacing}{\spaceskip=0pt\relax}
\providecommand{\BIBentryALTinterwordstretchfactor}{4}
\providecommand{\BIBentryALTinterwordspacing}{\spaceskip=\fontdimen2\font plus
\BIBentryALTinterwordstretchfactor\fontdimen3\font minus
  \fontdimen4\font\relax}
\providecommand{\BIBforeignlanguage}[2]{{%
\expandafter\ifx\csname l@#1\endcsname\relax
\typeout{** WARNING: IEEEtran.bst: No hyphenation pattern has been}%
\typeout{** loaded for the language `#1'. Using the pattern for}%
\typeout{** the default language instead.}%
\else
\language=\csname l@#1\endcsname
\fi
#2}}
\providecommand{\BIBdecl}{\relax}
\BIBdecl

\bibitem{Vapnik2013nature}
V.~Vapnik, \emph{The Nature of Statistical Learning Theory}, 2nd~ed.\hskip 1em
  plus 0.5em minus 0.4em\relax New York, NY: Springer-Verlag, 1999.

\bibitem{sebastiani2002machine}
F.~Sebastiani, ``Machine learning in automated text categorization,'' \emph{ACM
  Computing Surveys}, vol.~34, no.~1, pp. 1--47, 2002.

\bibitem{kotsiantis2007supervised}
S.~B. Kotsiantis, I.~Zaharakis, and P.~Pintelas, ``Supervised machine learning:
  {A} review of classification techniques,'' \emph{Emerging Artificial Intell.
  Applicat. Comput. Eng.}, vol. 160, pp. 3--24, 2007.

\bibitem{bengio2009learning}
Y.~Bengio, ``Learning deep architectures for {AI},'' \emph{Found. and Trends
  Mach. Learning}, vol.~2, no.~1, pp. 1--127, 2009.

\bibitem{Mohri2012foundations}
M.~Mohri, A.~Rostamizadeh, and A.~Talwalkar, \emph{Foundations of Machine
  Learning}, 2nd~ed.\hskip 1em plus 0.5em minus 0.4em\relax Cambridge, MA: MIT
  Press, 2018.

\bibitem{MLtextbook}
R.~M. Golden, \emph{Statistical Machine Learning: A Unified Framework}.\hskip
  1em plus 0.5em minus 0.4em\relax Boca Raton, FL: Chapman and Hall/CRC, 2020.

\bibitem{YangGangEtAl.ISPM20}
Z.~Yang, A.~Gang, and W.~U. Bajwa, ``Adversary-resilient distributed and
  decentralized statistical inference and machine learning: {{An}} overview of
  recent advances under the {{Byzantine}} threat model,'' \emph{IEEE Signal
  Process. Mag.}, vol.~37, no.~3, pp. 146--159, May 2020.

\bibitem{nokleby2020}
M.~{Nokleby}, H.~{Raja}, and W.~U. {Bajwa}, ``Scaling-up distributed processing
  of data streams for machine learning,'' \emph{Proceedings of the IEEE}, vol.
  108, no.~11, pp. 1984--2012, 2020.

\bibitem{predd2006distributed}
J.~B. Predd, S.~B. Kulkarni, and H.~V. Poor, ``Distributed learning in wireless
  sensor networks,'' \emph{IEEE Signal Process. Mag.}, vol.~23, no.~4, pp.
  56--69, 2006.

\bibitem{boyd2011distributed}
S.~Boyd, N.~Parikh, E.~Chu, B.~Peleato, and J.~Eckstein, ``Distributed
  optimization and statistical learning via the alternating direction method of
  multipliers,'' \emph{Found. and Trends Mach. Learning}, vol.~3, no.~1, pp.
  1--122, 2011.

\bibitem{AliH.Sayed2014}
A.~H. Sayed, ``Adaptation, learning, and optimization over networks,''
  \emph{Found. and Trends Mach. Learning}, vol.~7, no. 4-5, pp. 311--801, 2014.

\bibitem{nedic2018}
A.~Nedić, A.~Olshevsky, and M.~G. Rabbat, ``Network topology and
  communication-computation tradeoffs in decentralized optimization,''
  \emph{Proceedings of the IEEE}, vol. 106, no.~5, pp. 953--976, 2018.

\bibitem{sun2021decentralized}
T.~Sun, D.~Li, and B.~Wang, ``Decentralized federated averaging,'' \emph{arXiv
  preprint arXiv:2104.11375}, 2021.

\bibitem{Driscoll2003byzantine}
K.~Driscoll, B.~Hall, H.~Sivencrona, and P.~Zumsteq, ``Byzantine fault
  tolerance, from theory to reality,'' in \emph{Proc. Int. Conf. Computer
  Safety, Reliability, and Security (SAFECOMP'03)}, 2003, pp. 235--248.

\bibitem{Su2016fault}
L.~Su and N.~H. Vaidya, ``Fault-tolerant multi-agent optimization: {O}ptimal
  iterative distributed algorithms,'' in \emph{Proc. ACM Symp. Principles of
  Distributed Computing}, 2016, pp. 425--434.

\bibitem{Lamport1982byzantine}
L.~Lamport, R.~Shostak, and M.~Pease, ``The {B}yzantine generals problem,''
  \emph{ACM Trans. Programming Languages and Syst.}, vol.~4, no.~3, pp.
  382--401, 1982.

\bibitem{dutta2005best}
P.~Dutta, R.~Guerraoui, and M.~Vukolic, ``Best-case complexity of asynchronous
  {B}yzantine consensus,'' EPFL/IC/200499, Tech. Rep., 2005.

\bibitem{sousa2012byzantine}
J.~Sousa and A.~Bessani, ``From {B}yzantine consensus to {BFT} state machine
  replication: {A} latency-optimal transformation,'' in \emph{Proc. 9th Euro.
  Dependable Computing Conf.(EDCC'12)}, 2012, pp. 37--48.

\bibitem{Muli2014}
M.~Li, D.~G. Andersen, J.~W. Park, A.~J. Smola, A.~Ahmed, V.~Josifovski,
  J.~Long, E.~J. Shekita, and B.-Y. Su, ``Scaling distributed machine learning
  with the parameter server,'' in \emph{Proc. 11th {USENIX} Symp. Operating
  Systems Design and Implementation ({OSDI}'14)}, Broomfield, CO, Oct. 2014,
  pp. 583--598.

\bibitem{jakub2016}
J.~Konečný, H.~B. McMahan, F.~X. Yu, P.~Richtarik, A.~T. Suresh, and
  D.~Bacon, ``Federated learning: Strategies for improving communication
  efficiency,'' in \emph{Proc. NeurIPS Workshop on Private Multi-Party Machine
  Learning}, 2016.

\bibitem{blanchard2017machine}
P.~Blanchard, R.~Guerraoui, and J.~Stainer, ``Machine learning with
  adversaries: {B}yzantine tolerant gradient descent,'' in \emph{Proc. Advances
  in Neural Inf. Process. Syst.}, 2017, pp. 118--128.

\bibitem{draco2017chen}
L.~Chen, H.~Wang, Z.~Charles, and D.~Papailiopoulos, ``{DRACO}:
  {B}yzantine-resilient distributed training via redundant gradients,'' in
  \emph{Proc. 35th Intl. Conf. Machine Learning (ICML)}, 2018, pp. 903--912.

\bibitem{cao2018robust}
X.~Cao and L.~Lai, ``Robust distributed gradient descent with arbitrary number
  of {B}yzantine attackers,'' in \emph{Proc. IEEE Int. Conf. Acoust. Speech and
  Signal Process. (ICASSP'19)}, 2018, pp. 6373--6377.

\bibitem{su2018securing}
L.~Su and J.~Xu, ``Securing distributed machine learning in high dimensions,''
  \emph{arXiv preprint arXiv:1804.10140}, 2018.

\bibitem{yin2018defending}
D.~Yin, Y.~Chen, R.~Kannan, and P.~Bartlett, ``Defending against saddle point
  attack in {B}yzantine-robust distributed learning,'' in \emph{Proc. 36th
  Intl. Conf. Machine Learning}, Jun. 2019, pp. 7074--7084.

\bibitem{damaskinos2018asynchronous}
G.~Damaskinos, E.~E. Mhamdi, R.~Guerraoui, R.~Patra, and M.~Taziki,
  ``Asynchronous {B}yzantine machine learning (the case of {SGD}),'' in
  \emph{Proc. 35th Int. Conf. Machine Learning}, 2018, pp. 1145--1154.

\bibitem{mhamdi2018hidden}
E.~E. Mhamdi, R.~Guerraoui, and S.~Rouault, ``The hidden vulnerability of
  distributed learning in {B}yzantium,'' in \emph{Proc. 35th Int. Conf. Machine
  Learning}, 2018, pp. 3521--3530.

\bibitem{yin2018byzantine}
D.~Yin, Y.~Chen, K.~Ramchandran, and P.~Bartlett, ``Byzantine-robust
  distributed learning: {T}owards optimal statistical rates,'' in \emph{Proc.
  35th Intl. Conf. Machine Learning}, Jul. 2018, pp. 5650--5659.

\bibitem{alistarh2018byzantine}
D.~Alistarh, Z.~Allen-Zhu, and J.~Li, ``Byzantine stochastic gradient
  descent,'' in \emph{Proc. Advances in Neural Information Processing Systems},
  2018, pp. 4618--4628.

\bibitem{xie2018zeno}
C.~Xie, O.~Koyejo, and I.~Gupta, ``Zeno: {B}yzantine-suspicious stochastic
  gradient descent,'' \emph{arXiv preprint arXiv:1805.10032}, 2018.

\bibitem{xie2018generalized}
------, ``Generalized {B}yzantine-tolerant {SGD},'' \emph{arXiv preprint
  arXiv:1802.10116}, 2018.

\bibitem{xie2018phocas}
------, ``Phocas: {D}imensional {B}yzantine-resilient stochastic gradient
  descent,'' \emph{arXiv preprint arXiv:1805.09682}, 2018.

\bibitem{chen2019distributed}
X.~Chen, T.~Chen, H.~Sun, S.~Wu, and M.~Hong, ``Distributed training with
  heterogeneous data: {B}ridging median- and mean-based algorithms,'' in
  \emph{Proc. Advances in Neural Information Processing Systems}, 2020, pp.
  21\,616--21\,626.

\bibitem{rajput2019detox}
S.~Rajput, H.~Wang, Z.~Charles, and D.~Papailiopoulos, ``{DETOX}: {A}
  redundancy-based framework for faster and more robust gradient aggregation,''
  in \emph{Proc. Advances in Neural Information Processing Systems}, 2019.

\bibitem{li2018rsa}
L.~Li, W.~Xu, T.~Chen, G.~Giannakis, and Q.~Ling, ``{RSA}: {B}yzantine-robust
  stochastic aggregation methods for distributed learning from heterogeneous
  datasets,'' in \emph{Proc. AAAI Conference on Artificial Intelligence},
  vol.~33, 2019, pp. 1544--1551.

\bibitem{jin2019distributed}
R.~Jin, X.~He, and H.~Dai, ``Distributed {B}yzantine tolerant stochastic
  gradient descent in the era of big data,'' in \emph{Proc. IEEE Intl. Conf.
  Communications (ICC)}, 2019, pp. 1--6.

\bibitem{lin2019byzantine}
F.~Lin, Q.~Ling, and Z.~Xiong, ``Byzantine-resilient distributed large-scale
  matrix completion,'' in \emph{Proc. IEEE Int. Conf. Acoust. Speech and Signal
  Process. (ICASSP'19)}, 2019, pp. 8167--8171.

\bibitem{ghosh2019robust}
A.~Ghosh, J.~Hong, D.~Yin, and K.~Ramchandran, ``Robust federated learning in a
  heterogeneous environment,'' \emph{arXiv preprint arXiv:1906.06629}, 2019.

\bibitem{data2019data}
D.~Data, L.~Song, and S.~N. Diggavi, ``Data encoding for {B}yzantine-resilient
  distributed optimization,'' \emph{IEEE Transactions on Information Theory},
  vol.~67, no.~2, pp. 1117--1140, 2021.

\bibitem{el2019sgd}
E.~M.~E. Mhamdi, R.~Guerraoui, A.~Guirguis, and S.~Rouault, ``{SGD:}
  {D}ecentralized {B}yzantine resilience,'' \emph{arXiv preprint
  arXiv:1905.03853}, 2019.

\bibitem{xie2019zeno++}
C.~Xie, S.~Koyejo, and I.~Gupta, ``Zeno++: {R}obust fully asynchronous {SGD},''
  in \emph{Proc. 37th Intl. Conf. Machine Learning}, Jul. 2020, pp.
  10\,495--10\,503.

\bibitem{elmhamdi2019fast}
E.-M. {El-Mhamdi}, R.~{Guerraoui}, and S.~{Rouault}, ``Fast and robust
  distributed learning in high dimension,'' \emph{arXiv preprint
  arXiv:1905.04374}, 2019.

\bibitem{nedic2009}
A.~Nedic and A.~Ozdaglar, ``Distributed subgradient methods for multi-agent
  optimization,'' \emph{IEEE Trans. Autom. Control}, vol.~54, no.~1, pp.
  48--61, 2009.

\bibitem{ram2010distributed}
S.~S. Ram, A.~Nedi{\'{c}}, and V.~Veeravalli, ``Distributed stochastic
  subgradient projection algorithms for convex optimization,'' \emph{J. Optim.
  Theory and Appl.}, vol. 147, no.~3, pp. 516--545, 2010.

\bibitem{nedic2015distributed}
A.~Nedi{\'{c}} and A.~Olshevsky, ``Distributed optimization over time-varying
  directed graphs,'' \emph{IEEE Trans. Autom. Control}, vol.~60, no.~3, pp.
  601--615, 2015.

\bibitem{nedic2020}
S.~Pu and A.~Nedić, ``Distributed stochastic gradient tracking methods,''
  \emph{Mathematical Programming}, vol. 187, pp. 409--457, 2021.

\bibitem{forero2010consensus}
P.~A. Forero, A.~Cano, and G.~B. Giannakis, ``Consensus-based distributed
  support vector machines,'' \emph{J. Mach. Learning Research}, vol.~11, pp.
  1663--1707, 2010.

\bibitem{mota2013admm}
J.~F. Mota, J.~M. Xavier, P.~M. Aquiar, and M.~Puschel, ``D-{ADMM}: A
  communication-efficient distributed algorithm for separable optimization,''
  \emph{IEEE Trans. Signal Process.}, vol.~61, no.~10, pp. 2718--2723, 2013.

\bibitem{shi2014on}
W.~Shi, Q.~Ling, K.~Yuan, G.~Wu, and W.~Yin, ``On the linear convergence of the
  {ADMM} in decentralized consensus optimization,'' \emph{IEEE Trans. Signal
  Process.}, vol.~62, no.~7, pp. 1750--1761, 2014.

\bibitem{Mokhtari2016decentralized}
A.~Mokhtari, W.~Shi, Q.~Ling, and A.~Ribeiro, ``A decentralized second-order
  method with exact linear convergence rate for consensus optimization,''
  \emph{IEEE Trans. Signal Inf. Process. Netw.}, vol.~2, no.~4, pp. 507--522,
  2016.

\bibitem{Mokhtari2017network}
A.~Mokhtari, Q.~Ling, and A.~Ribeiro, ``Network {N}ewton distributed
  optimization methods,'' \emph{IEEE Trans. Signal Process.}, vol.~65, no.~1,
  pp. 146--161, 2017.

\bibitem{leblanc2013resilient}
H.~J. LeBlanc, H.~Zhang, X.~Koutsoukos, and S.~Sundaram, ``Resilient asymptotic
  consensus in robust networks,'' \emph{IEEE J. Sel. Areas in Commun.},
  vol.~31, no.~4, pp. 766--781, 2013.

\bibitem{vaidya2014iterative}
N.~H. Vaidya, L.~Tseng, and G.~Liang, ``Iterative {B}yzantine vector consensus
  in incomplete graphs,'' in \emph{Proc. 15th Int. Conf. Distributed Computing
  and Networking}, 2014, pp. 14--28.

\bibitem{sundaram2018distributed}
S.~Sundaram and B.~Gharesifard, ``Distributed optimization under adversarial
  nodes,'' \emph{IEEE Trans. Autom. Control}, vol.~64, no.~3, pp. 1063--1076,
  2019.

\bibitem{yang2016rdsvm}
Z.~Yang and W.~U. Bajwa, ``{RD-SVM}: {A} resilient distributed support vector
  machine,'' in \emph{Proc. IEEE Int. Conf. Acoust. Speech and Signal Process.
  (ICASSP'16)}, 2016, pp. 2444--2448.

\bibitem{xu2018robust}
W.~Xu, Z.~Li, and Q.~Ling, ``Robust decentralized dynamic optimization at
  presence of malfunctioning agents,'' \emph{Signal Process.}, vol. 153, pp.
  24--33, 2018.

\bibitem{mitra2019resilient}
A.~Mitra, J.~Richards, S.~Bagchi, and S.~Sundaram, ``Resilient distributed
  state estimation with mobile agents: {O}vercoming {B}yzantine adversaries,
  communication losses, and intermittent measurements,'' \emph{Autonomous
  Robots}, vol.~43, no.~3, pp. 743--768, 2019.

\bibitem{su2018finite}
L.~Su and S.~Shahrampour, ``Finite-time guarantees for {B}yzantine-resilient
  distributed state estimation with noisy measurements,'' \emph{IEEE
  Transactions on Automatic Control}, vol.~65, no.~9, pp. 3758--3771, 2020.

\bibitem{yang2019byrdie}
Z.~Yang and W.~U. Bajwa, ``{{ByRDiE}}: Byzantine-resilient distributed
  coordinate descent for decentralized learning,'' \emph{IEEE Trans. Signal
  Inf. Process. Netw.}, vol.~5, no.~4, pp. 611--627, Dec. 2019.

\bibitem{Kuwaranancharoen2020ByzantineResilientDO}
K.~Kuwaranancharoen, L.~Xin, and S.~Sundaram, ``Byzantine-resilient distributed
  optimization of multi-dimensional functions,'' in \emph{Proc. American
  Control Conference (ACC)}, 2020, pp. 4399--4404.

\bibitem{Peng2020ByzantineRobustDS}
J.~Peng, W.~Li, and Q.~Ling, ``Byzantine-robust decentralized stochastic
  optimization over static and time-varying networks,'' \emph{Signal
  Processing}, vol. 183, p. 108020, 2021.

\bibitem{Guo2020TowardsBL}
S.~Guo, T.~Zhang, X.~Xie, L.~Ma, T.~Xiang, and Y.~Liu, ``Towards
  {B}yzantine-resilient learning in decentralized systems,'' \emph{arXiv
  preprint arXiv:2002.08569}, 2020.

\bibitem{ElMhamdi2020CollaborativeLA}
E.-M. El-Mhamdi, R.~Guerraoui, A.~Guirguis, L.~Hoang, and S.~Rouault,
  ``Collaborative learning as an agreement problem,'' \emph{arXiv preprint
  arXiv:2008.00742v3}, 2020.

\bibitem{Lorenzo2016NEXTIN}
P.~D. Lorenzo and G.~Scutari, ``Next: In-network nonconvex optimization,''
  \emph{IEEE Transactions on Signal and Information Processing over Networks},
  vol.~2, pp. 120--136, 2016.

\bibitem{Zeng2018OnND}
J.~Zeng and W.~Yin, ``On nonconvex decentralized gradient descent,'' \emph{IEEE
  Transactions on Signal Processing}, vol.~66, pp. 2834--2848, 2018.

\bibitem{Sun2019ImprovingTS}
H.~Sun, S.~Lu, and M.~Hong, ``Improving the sample and communication complexity
  for decentralized non-convex optimization: Joint gradient estimation and
  tracking,'' in \emph{Proc. 37th Intl. Conf. Machine Learning}, Jul. 2020, pp.
  9217--9228.

\bibitem{xin2021fast}
R.~{Xin}, U.~A. {Khan}, and S.~{Kar}, ``{Fast decentralized non-convex
  finite-sum optimization with recursive variance reduction},'' \emph{arXiv
  preprint arXiv:2008.07428}, 2020.

\bibitem{Lie2022Byzantine}
L.~He, S.~P. Karimireddy, and M.~Jaggi, ``Byzantine-robust decentralized
  learning via self-centered clipping,'' \emph{arXiv preprint
  arXiv:2202.01545}, 2022.

\bibitem{Lecun1998}
Y.~Lecun, L.~Bottou, Y.~Bengio, and P.~Haffner, ``Gradient-based learning
  applied to document recognition,'' \emph{Proceedings of the IEEE}, vol.~86,
  no.~11, pp. 2278--2324, 1998.

\bibitem{krizhevsky2009learning}
A.~Krizhevsky and G.~Hinton, ``Learning multiple layers of features from tiny
  images,'' 2009.

\bibitem{sohrab2003basic}
H.~H. Sohrab, \emph{Basic Real Analysis}, 2nd~ed.\hskip 1em plus 0.5em minus
  0.4em\relax New York, NY: Springer, 2003.

\bibitem{Sun2015nonconvex}
J.~Sun, Q.~Qu, and J.~Wright, ``When are nonconvex problems not scary?''
  \emph{arXiv preprint arXiv:1510.06096}, 2015.

\bibitem{Jain2017nonconvex}
P.~Jain and P.~Kar, ``Non-convex optimization for machine learning,''
  \emph{Foundations and Trends in Machine Learning}, vol.~10, no. 3-4, p.
  142–336, 2017.

\bibitem{Yonel2020}
B.~Yonel and B.~Yazici, ``A deterministic theory for exact non-convex phase
  retrieval,'' \emph{IEEE Transactions on Signal Processing}, vol.~68, pp.
  4612--4626, 2020.

\bibitem{zhou2016noncovex}
Y.~Zhou, H.~Zhang, and Y.~Liang, ``Geometrical properties and accelerated
  gradient solvers of non-convex phase retrieval,'' in \emph{Proc. 54th Annu.
  Allerton Conf. Communication, Control, and Computing}, 2016, pp. 331--335.

\bibitem{ypma1984local}
T.~Ypma, ``Local convergence of inexact {N}ewton methods,'' \emph{SIAM Journal
  on Numerical Analysis}, vol.~21, no.~3, pp. 583--590, 1984.

\bibitem{ochs2018local}
P.~Ochs, ``Local convergence of the heavy-ball method and i{P}iano for
  non-convex optimization,'' \emph{Journal of Optimization Theory and
  Applications}, vol. 177, no.~1, pp. 153--180, 2018.

\bibitem{bock2019proof}
S.~Bock and M.~Wei{\ss}, ``A proof of local convergence for the {A}dam
  optimizer,'' in \emph{Proc. International Joint Conference on Neural Networks
  (IJCNN)}.\hskip 1em plus 0.5em minus 0.4em\relax IEEE, 2019, pp. 1--8.

\bibitem{duchi2012dual}
J.~C. Duchi, A.~Agarwal, and M.~J. Wainwright, ``Dual averaging for distributed
  optimization: Convergence analysis and network scaling,'' \emph{IEEE Trans.
  Autom. control}, vol.~57, no.~3, pp. 592--606, 2012.

\bibitem{su2015byzantine}
L.~Su and N.~Vaidya, ``Multi-agent optimization in the presence of {B}yzantine
  adversaries: Fundamental limits,'' in \emph{Proc. American Control Conference
  (ACC)}, 2016, pp. 7183--7188.

\bibitem{Yang2019ByzantineResilientSG}
H.~Yang, X.~zhong Zhang, M.~Fang, and J.~Liu, ``Byzantine-resilient stochastic
  gradient descent for distributed learning: A {L}ipschitz-inspired
  coordinate-wise median approach,'' \emph{Proc. IEEE Conference on Decision
  and Control (CDC)}, pp. 5832--5837, 2019.

\bibitem{Data2020ByzantineResilientSI}
D.~Data and S.~Diggavi, ``Byzantine-resilient high-dimensional {SGD} with local
  iterations on heterogeneous data,'' in \emph{Proc. 38th Intl. Conf. Machine
  Learning}, Jul. 2021, pp. 2478--2488.

\bibitem{huber2011robust}
P.~J. Huber, \emph{Robust Statistics}.\hskip 1em plus 0.5em minus 0.4em\relax
  Berlin, Heidelberg: Springer, 2011.

\bibitem{Su2015fault}
L.~Su and N.~H. Vaidya, ``Fault-tolerant distributed optimization (part {IV):}
  {C}onstrained optimization with arbitrary directed networks,'' \emph{arXiv
  preprint arXiv:1511.01821}, 2015.

\bibitem{Nesterovbook}
Y.~Nesterov, \emph{Introductory Lectures on Convex Optimization}, ser. Applied
  optimization; v. 87.\hskip 1em plus 0.5em minus 0.4em\relax Springer US,
  2004.

\bibitem{yang2019bridge}
\BIBentryALTinterwordspacing
Z.~Yang and W.~U. Bajwa, ``{BRIDGE}: {B}yzantine-resilient decentralized
  gradient descent,'' \emph{arXiv preprint arXiv:1908.08098v1}, Aug. 2019.
  [Online]. Available: \url{https://arxiv.org/abs/1908.08098v1}
\BIBentrySTDinterwordspacing

\bibitem{Song2018}
S.~Mei, Y.~Bai, and A.~Montanari, ``{The landscape of empirical risk for
  nonconvex losses},'' \emph{The Annals of Statistics}, vol.~46, no.~6A, pp.
  2747 -- 2774, 2018.

\bibitem{Damek2021}
D.~Davis and D.~Drusvyatskiy, ``Graphical convergence of subgradients in
  nonconvex optimization and learning,'' \emph{Mathematics of Operations
  Research}, vol.~47, no.~1, pp. 209--231, 2021.

\bibitem{Foster2018}
D.~J. Foster, A.~Sekhari, and K.~Sridharan, ``Uniform convergence of gradients
  for non-convex learning and optimization,'' in \emph{Proc. Advances in Neural
  Information Processing Systems}, vol.~31.\hskip 1em plus 0.5em minus
  0.4em\relax Curran Associates, Inc., 2018.

\bibitem{le2016}
\BIBentryALTinterwordspacing
T.~V.~J. Le and N.~Gopee, ``Classifying {CIFAR}-10 images using unsupervised
  feature \& ensemble learning,'' Dec 2016. [Online]. Available:
  \url{https://trucvietle.me/files/601-report.pdf}
\BIBentrySTDinterwordspacing

\bibitem{hoeffding1963probability}
W.~Hoeffding, ``{P}robability inequalities for sums of bounded random
  variables,'' \emph{J. American Stat. Assoc.}, vol.~58, no. 301, pp. 13--30,
  1963.

\bibitem{Verger-Gaugry2005covering}
J.~Verger-Gaugry, ``Covering a ball with smaller equal balls in {$R^n$},''
  \emph{Discrete {\&} Computational Geometry}, vol.~33, no.~1, pp. 143--155,
  2005.

\end{thebibliography}
\end{document}